\documentclass[sigconf]{acmart}
\AtBeginDocument{%
  \providecommand\BibTeX{{%
    \normalfont B\kern-0.5em{\scshape i\kern-0.25em b}\kern-0.8em\TeX}}}

\acmConference[]{The Web Conference}{MAY 13 - 17, 2024}{Singapore}
\newcommand{\systemname}{{KwaiAgents}}

\usepackage{multirow}
\usepackage{color}
\usepackage{amsmath}
\usepackage{listings}
\usepackage{xcolor}

\colorlet{punct}{red!60!black}
\definecolor{background}{HTML}{EEEEEE}
\definecolor{delim}{RGB}{20,105,176}
\colorlet{numb}{magenta!60!black}

\lstdefinelanguage{json}{
    basicstyle=\normalfont\ttfamily,
    numbers=left,
    numberstyle=\scriptsize,
    stepnumber=1,
    numbersep=8pt,
    showstringspaces=false,
    breaklines=true,
    frame=lines,
    backgroundcolor=\color{background},
    literate=
     *{0}{{{\color{numb}0}}}{1}
      {1}{{{\color{numb}1}}}{1}
      {2}{{{\color{numb}2}}}{1}
      {3}{{{\color{numb}3}}}{1}
      {4}{{{\color{numb}4}}}{1}
      {5}{{{\color{numb}5}}}{1}
      {6}{{{\color{numb}6}}}{1}
      {7}{{{\color{numb}7}}}{1}
      {8}{{{\color{numb}8}}}{1}
      {9}{{{\color{numb}9}}}{1}
      {:}{{{\color{punct}{:}}}}{1}
      {,}{{{\color{punct}{,}}}}{1}
      {\{}{{{\color{delim}{\{}}}}{1}
      {\}}{{{\color{delim}{\}}}}}{1}
      {[}{{{\color{delim}{[}}}}{1}
      {]}{{{\color{delim}{]}}}}{1},
}

\begin{document}

\title{\systemname: Generalized Information-seeking Agent System with Large Language Models}


% \author{Anonymous authors}

% \affiliation{~}
\author{Haojie Pan$^1$, Zepeng Zhai$^1$, Hao Yuan, Yaojia Lv$^2$, Ruiji Fu$^{1}$, Ming Liu$^2$, \\ Zhongyuan Wang$^1$, Bing Qin$^2$}

\affiliation{$^1$ Kuaishou Inc. $^2$ Harbin Institute of Technology \\
\institution{\{panhaojie,zhaizepeng03,furuiji,wangzhongyuan\}@kuaishou.com}
\country{\{yuanhaov\}@gmail.com, \{yjlv, mliu, qinb\}@ir.hit.edu.cn}
\city{~ \\ ~ \\ ~ \\ ~}
\vspace{-9pt}
}


\begin{abstract}
  % A clear and well-documented \LaTeX\ document is presented as an
  % article formatted for publication by ACM in a conference proceedings
  % or journal publication. Based on the ``acmart'' document class, this
  % article presents and explains many of the common variations, as well
  % as many of the formatting elements an author may use in the
  % preparation of the documentation of their work.

Driven by curiosity, humans have continually sought to explore and understand the world around them, leading to the invention of various tools to satiate this inquisitiveness. Despite not having the capacity to process and memorize vast amounts of information in their brains, humans excel in critical thinking, planning, reflection, and harnessing available tools to interact with and interpret the world, enabling them to find answers efficiently.
The recent advancements in large language models (LLMs) suggest that machines might also possess the aforementioned human-like capabilities, allowing them to exhibit powerful abilities even with a constrained parameter count.
In this paper, we introduce \systemname, a generalized information-seeking agent system based on LLMs. Within \systemname, we propose an agent system that employs LLMs as its cognitive core, which is capable of understanding a user's query, behavior guidelines, and referencing external documents. The agent can also update and retrieve information from its internal memory, plan and execute actions using a time-aware search-browse toolkit, and ultimately provide a comprehensive response. We further investigate the system's performance when powered by LLMs less advanced than GPT-4, and introduce the Meta-Agent Tuning (MAT) framework, designed to ensure even an open-sourced 7B or 13B model performs well among many agent systems. We exploit both benchmark and human evaluations to systematically validate these capabilities. Extensive experiments show the superiority of our agent system compared to other autonomous agents and highlight the enhanced generalized agent-abilities of our fine-tuned LLMs.
\footnote{We release a lite version of the system code, models, training data, and benchmarks for public use on \href{https://github.com/KwaiKEG/KwaiAgents}{https://github.com/KwaiKEG/KwaiAgents}.}

\end{abstract}

\maketitle

\section{Introduction} \label{intro}

Bertrand Russell once profoundly stated that the search for knowledge is one of the simple but overwhelmingly strong passions that have governed his life~\citep{russell-passion}. 
Through the ages, successive generations have committed themselves to probing the intricacies of our world, crafting ingenious tools for organization and retrieval, all in a bid to quench insatiable curiosity. 
However, Research in cognitive science reveals that humans, on average, forget approximately 50\% of newly acquired information within an hour, a phenomenon termed the forgetting curve~\citep{forgetting-curve}. 
This observation holds particularly true when individuals attempt to internalize knowledge without periodic reinforcement. For instance, while many can effortlessly recall that Mount Everest is ``the highest mountain in the world'', the identity of ``the fifth highest mountain'' often eludes memory. 
However, the human forte lies in critical thinking, planning, reflection, and the adept use of external resources. Confronted with a knowledge gap, individuals might consult search engines like Google, or turn to repositories of knowledge like Wikipedia or books. This unique blend of cognition and resourcefulness distinguishes humans from other species, often rendering us more insightful than even the most advanced computers.

\begin{figure*}[t]
\centering
\includegraphics[height=0.55\textwidth]{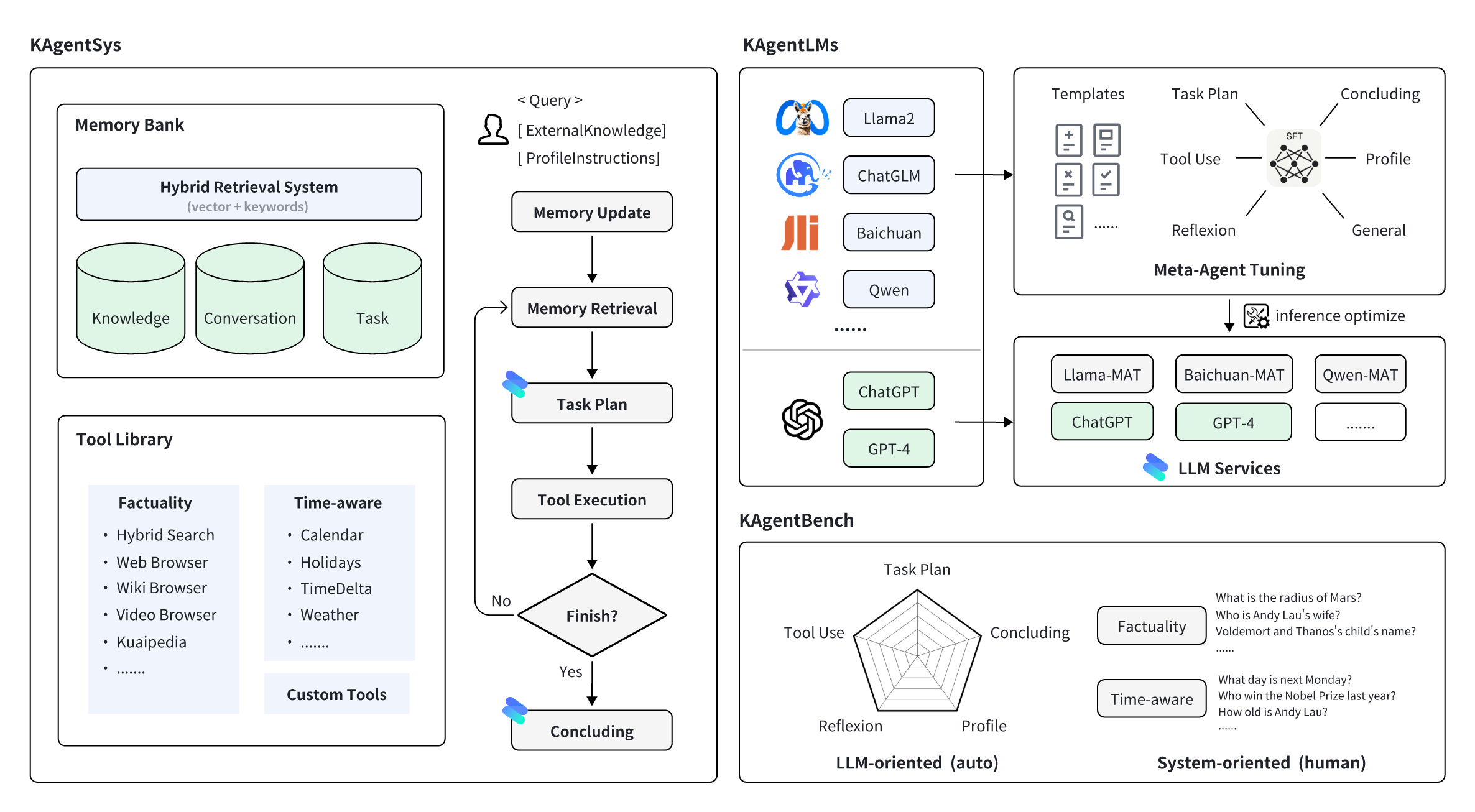}
\caption{The overview of \systemname, which contain three components: system, LLMs, and benchmark.}\label{fig:overview}
\end{figure*}

Recent advances in Large Language Models (LLMs) are increasingly seen as catalysts for the development of Artificial General Intelligence (AGI). Research indicates that LLMs are capable of following human instructions and exhibit skills in diverse areas including conversation, mathematics, coding, comprehension, and commonsense reasoning~\citep{ouyang-instruct-tuning, gpt4-tech-report, Emergent-abilities-of-llms}.
Beyond these abilities, recent studies highlight LLMs' proficiency in planning, reflection, and the use of external tools~\citep{yao-react,aman-self-refine,shinn2023reflexion,timo-toolformer}.
This has led to the emergence of a new paradigm in AI, where agents powered by LLMs are being developed~\citep{lilianweng-autoagents,norman-unified-agent,xi2023rise,wang2023survey}.
Closed-source LLMs, like ChatGPT and GPT-4, have demonstrated their utility in various agent systems. 
HuggingGPT~\citep{shen-hugginggpt} illustrates LLMs' ability to utilize AI-powered tools, while projects such as Auto-GPT~\citep{auto-gpt} and BabyAGI~\citep{baby-agi} show that LLM-powered agents can autonomously fulfill human requirements. Generative Agents~\citep{generative-agents} mimic human behaviors and interact within immersive communities. Voyager~\citep{wang2023voyager} demonstrates that agents can acquire diverse skills and continuously explore environments like Minecraft. AutoGen~\citep{wu2023autogen} and ChatDev~\citep{qian2023communicative} enable communication between agents for application development.
 On the other hand, open-source smaller models like Llama (7B or 13B)~\citep{touvron2023llama,touvron2023llama2} have shown their potential in specific agent
systems when fine-tuned with targeted instructional prompts~\cite{patil2023gorilla,qin2023toolllm,li2023modelscopeagent,zeng2023agenttuning}. However, these models' performance declines if prompt templates are altered. Thus it remains uncertain whether these smaller, open-source models have acquired generalizable agent capabilities or merely overfit the prompt templates.
In the context of the information-seeking scenario presented in this paper, many studies have also investigated how LLMs address web navigation challenges, encompassing tasks such as searching and retrieval~\citep{webgpt}, online shopping~\citep{yao2022webshop}, and interacting with webpages to accomplish specific objectives~\citep{deng2023mind2web,izzeddin23webagent}. However, information retrieved from web searches can be skewed by trendiness or be misleading and outdated. It's essential for LLMs to consider and reconcile information from multiple sources, especially when dealing with high-quality factual information, such as that found on Wikipedia, or how-to knowledge in videos~\citep{kuaipedia}.

% In this paper, we proposed \systemname, a comprehensive and generalized information-seeking agent system with LLMs. As shown in Figure \ref{fig:overview}, \systemname~ contains three parts: (1) LLMs: a set of open-sourced LLMs models continually tuned for the abilities of agents. (2) System: a typical autonomous agent loop incorporated with memory bank, tool library, task planner and a concluding module. (3) Benchmark: which takes into two part, the first is to evaluate how LLMs performs on different agent-system's prompts with different abilities, and the second is a set of time-aware or factual questions to evaluate whether agent system accompanied by LLMs can complete information-seeking tasks.

In this paper, we introduce \systemname, a generalized information-seeking agent system leveraging with LLMs. As illustrated in Figure \ref{fig:overview}, \systemname~comprises three main components: (1) \textbf{KAgentSys}, an autonomous agent loop that integrates a memory bank, a tool library, a task planner, and a concluding module. (2) \textbf{KAgentLMs}, which are a suite of open-source LLMs continuously fine-tuned to enhance agent capabilities.  (3) \textbf{KAgentBench}, a benchmark that assesses the performance of LLMs in responding to varied agent-system prompts across different capabilities; We also collect a series of fact-based or time-aware queries designed to evaluate the effectiveness of the LLM-enhanced agent system in executing information-seeking tasks.

Within the KAgentSys, we design a planning-concluding procedure,  wherein the planning component facilitates inter-agent thinking, and the concluding segment is tailored for human interaction within conversational sessions. Furthermore, we propose an innovative hybrid search-browse toolkit, sourcing knowledge from search results, webpages, Wikipedia, and various aspects and videos found in Kuaipedia~\citep{kuaipedia}. This hybrid search toolkit, along with a time-aware toolkit, effectively tackles long-tail, deceptive and obsolete information on the Internet.

KAgentLMs are produced to explore whether small, open-sourced models (7B or 13B) can master skills such as planning, reflection, tool-use in various agent systems. We introduce the \textbf{Meta-Agent Tuning (MAT)} framework. 
This framework aims to create an advanced structure for agent planning and reasoning prompts.
We analyzed many popular agent systems, distilling the prompts into six key components: (1) system profile, (2) instructions or constraints, (3) tool specifications, (4) goal placement, (5) memory allocation, and (6) output format.
Subsequently, a meta-agent prompting mechanism generates comprehensive instructional prompt templates, which are then incorporated into an experimental agent loop for comparison with results from promising open-sourced templates. Less effective prompt templates are filtered through a scoring mechanism. During the experimental agent looping, GPT-4 is utilized to generate responses to a variety of inquiries from the approved templates, forming the agent instruction tuning dataset. This dataset covers a wide range of scenarios pertinent to information-seeking agent systems, including multi-turn conversations, multi-step task planning, reflection, tool-use, integration of external knowledge, profile adaptation, and adherence to human directives.

Given the complexity of many agent systems and benchmarks~\citep{qin2023toolllm,liu2023agentbench}—often requiring API-keys or sandbox environments—it becomes challenging to evaluate LLMs using a straightforward prompt-in-response-out mechanism. Thus, we introduce KAgentBench, designed for the automated and streamlined assessment of different agent abilities across different models. We also carefully collect approximately 200 extra factual-based or time-aware queries, where one can hire human annotators to gauge the performance of various LLMs on different agent systems. Our comprehensive experiments reveal that our system outperforms several open-sourced agent systems. Notably, following MAT, even a 7B or 13B model exhibits the generalized agent capabilities required for information-seeking tasks, regardless of the system employed.

To sum up, we make the following contributions:
\begin{itemize}
    \item We propose the KAgentSys, which integrates a planning-concluding procedure with a novel hybrid search-browse and time-aware toolkit, demonstrating superior performance over many existing open-sourced agent systems.
    \item We introduce KAgentLMs and the Meta-Agent Tuning framework to explore how smaller open-sourced LLMs can acquire generalized agent capabilities for information-seeking tasks.
    \item We develop an accessible benchmark KAgentBench for the both automatic evaluation of capabilities necessary for various agent systems and human evaluation of queries for a comprehensive assessment of agent system performance.
\end{itemize}

% Main contributions:

% \begin{enumerate}
%     \item High-quality Data, including training data and a benchmark.
%     \item Bunch of Models fine-tuned on our data.
%     \item A system of information-seeking agent system.
% \end{enumerate}

% As shown in Figure \ref{fig:overview}. 

\begin{figure*}[t]
\centering
\includegraphics[height=0.47\textwidth]{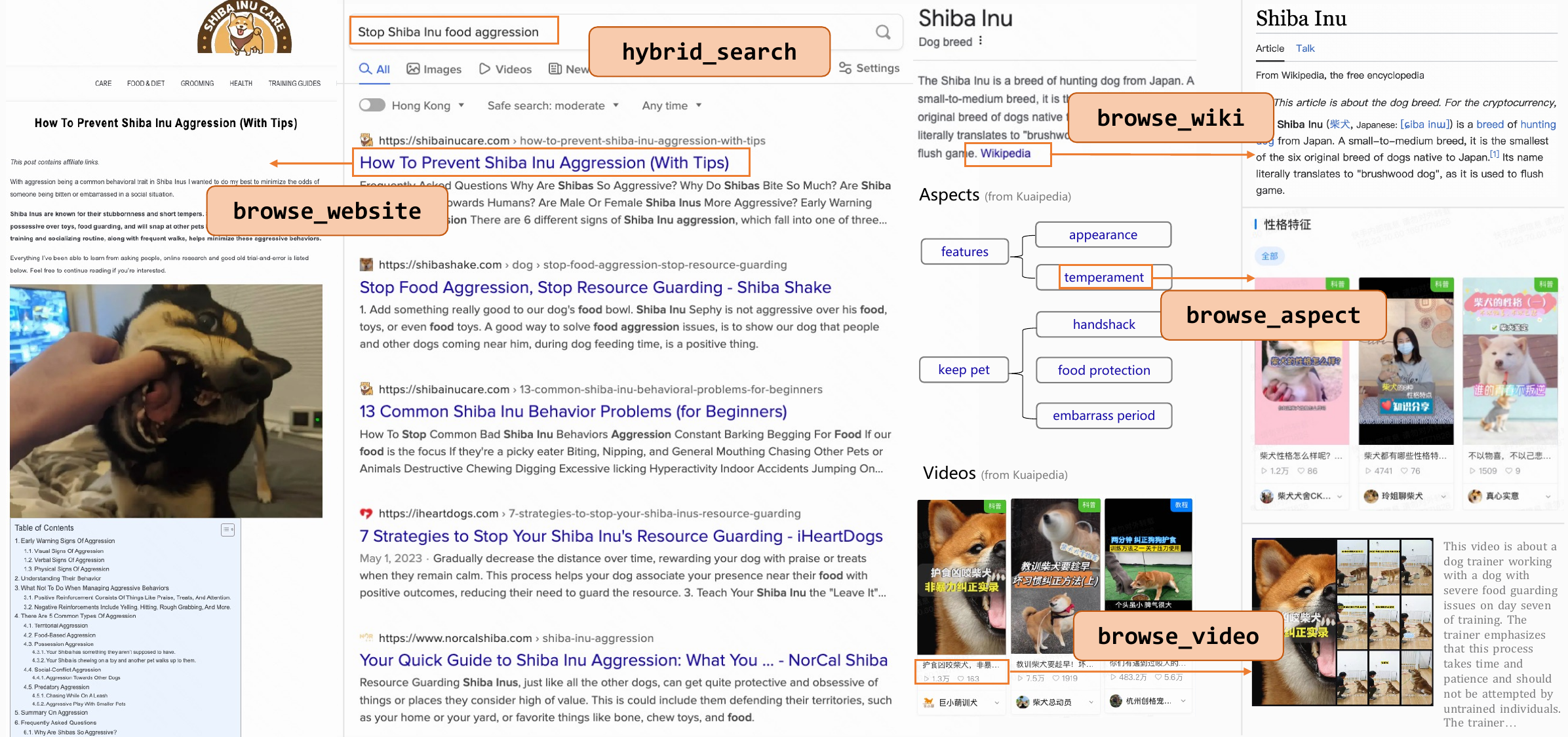}
\caption{An example of how \systemname use the hybrid toolset for information seeking.}\label{fig:fact-seeking-tools}
\end{figure*}

\section{System}
In this section, we delineate the functioning of KAgentSys ~and its response to user requirements. We begin by explaining the roles of LLMs, the memory bank, and the tool library. Subsequently, these components will be integrated into the primary agent loop of KAgentSys.

\subsection{LLMs}
In an analogy to the human brain, LLMs should first understand the user's requirements. This includes considering external knowledge pertinent to the current turn of the dialogue, which has been retrieved from a database, as well as referencing previous conversation messages and tasks completed in the past. Subsequently, the LLM should generate well-founded plans, appropriate tool commands, or draw conclusions upon completing the planning process. To accommodate the varying requirements of different LLMs, we introduce a straightforward calling API that accepts a prompt as input and returns a response.

\subsection{Memory Bank}
The memory module caches a user's contextual information throughout a conversation session. This information is categorized into three distinct components:

\noindent \textbf{Knowledge Memory}: This module captures and retrieves external resources that a user wishes to integrate into the conversational context. Examples include the user's personal data or documents they wish to discuss in detail.

\noindent \textbf{Conversation Memory}: This component records query-response pairs for every turn in the conversation.

\noindent \textbf{Task Memory}: This memory chronicles the decision-making process of the KAgentSys. For every conversation session, once the user inputs text, the KAgentSys strategize by planning tasks, selecting appropriate tools, and executing commands to obtain observations from these tools. Consequently, this task-related information is systematically stored in memory.

For each memory type, text is partitioned into segments with fixed-maximum-length. Each segment is then transformed into a vector representation for vector-based searches or indexed using an inverted list for keyword searches. This culminates in a hybrid retrieval system that, given a succinct query, extracts relevant text segments from the various memory types using different retrieval mechanisms. These segments are then aggregated and formatted into structured, context-rich text, making it ready for use in subsequent prompts. One can find more details and cases from appendix ~\ref{appendix-memory-bank} and ~\ref{memory-data}
% An example is shown in ??.

\subsection{Tool Library}

The tool library offers two distinct sets of pre-defined tools: one for factuality and the other for time-awareness.

Typical strategies, as outlined by ~\citep{webgpt,auto-gpt}, utilize search engines for fact retrieval by directing a LLM to generate a query. The LLM then determine whether to access a specific URL from the search results or respond to the query based on the retrieved information. If the LLM chooses to access a link, it proceeds to scan the webpage, extracting relevant paragraphs or generating concise summaries for the query.
Drawing inspiration from Google Search and it's Knowledge Graph~\citep{google-kg}, KAgentSys adopts the aforementioned search-browse mechanism. However, it extends this approach into a ``hybrid search'', which integrates traditional web searching with an entity search in Kuaipedia~\citep{kuaipedia}, as depicted in Figure \ref{fig:fact-seeking-tools}. The \texttt{hybrid\_search} function accepts a single argument, \texttt{query}, and concurrently performs two tasks:
(1) It employs a search engine API to fetch pertinent webpages, each accompanied by a title, URL, and a brief summary.
(2) It applies principal entity linking, as described in ~\citep{kuaipedia}, to identify the primary entity and gathers a concise Wikipedia description, the aspect tree, and the most relevant videos in Kuaipedia. Each component (entities, aspects, and videos) includes its corresponding URL for detailed exploration.
When the agent opts to browse, it selects one of the following actions:
(1) \texttt{browse\_website} to delve into a webpage and either summarize it or derive answers relevant to the input query.
(2) \texttt{browse\_wiki} to explore a Wikipedia page and either summarize it or extract answers pertinent to the query.
(3) \texttt{browse\_aspect} to examine the facets of an entity and investigate related videos.
(4) \texttt{browse\_video} to interpret the OCR of individual frames and the ASR of the video, producing a text-based summary.

The second suite of tools within KAgentSys emphasizes time-awareness. We observed that merely incorporating timestamps into prompts remains suboptimal for time-aware question-answering, leading to the propagation of outdated information. Therefore, we introduced several time-aware tools, including:
(1) \texttt{calendar}, which processes a date range to provide details from both the Gregorian and Lunar calendars.
(2) \texttt{holidays}, which, given a date range, returns notable festivals.
(3) \texttt{time\_delta}, which calculates the duration (in days/hours/minutes/seconds) between two timestamps.
(4) \texttt{weather}, which requires a location and date range to furnish weather details, such as temperature, precipitation, and humidity.

In addition, user-defined custom tools can also be seamlessly integrated into KAgentSys, provided they come with detailed specifications. The detailed format of all of the tools can be found in \ref{appendix-tool-library}.

\subsection{Agent Loop}
Before initializing the \systemname's loop, the user submits a query (required) alongside optional inputs such as external knowledge sources (link or file) and specific instructions or profiles that guide the agent's behavior. The loop then progresses through the following procedures:

\noindent \textbf{Memory Update}: The system updates records of the user's past interactions, including previous conversations and executed tasks. If the user provides new links or files, the external knowledge base is also updated accordingly.

\noindent \textbf{Memory Retrieval}: Based on the user's current query and recent interactions, relevant segments from external knowledge sources, past conversations, and completed tasks are retrieved. These segments are consolidated into a fixed-maximum-length text box for subsequent reference.

\noindent \textbf{Task Planning}: This phase leverages the gathered information to construct a comprehensive prompt for LLMs. The prompt encompasses the agent's profile, behavior guidelines as specified by the user, tool specifications, the consolidated memory text box, the current timestamp, and a predefined output format in JSON. Consequently, the LLM is prompted to generate a \texttt{task\_name}, representing the plan for the ensuing phase, and a \texttt{command} detailing which tools to invoke and the respective arguments.

\noindent \textbf{Tool Execution}: If the ``Task Planning'' phase yields a tool named \texttt{task\_complete}, the loop terminates, prompting the agent to formulate a conclusion. Otherwise, the specified commands are executed. Each command produces an observation in a custom format, which is then integrated into the task memory.

\noindent \textbf{Concluding}: Analogous to the ``Task Planning'' phase, the agent formulates a response to the user's query by considering the retrieved memories. The agent's profile, user-provided instructions, and the current timestamp are also incorporated into the final prompt.

Both of the prompt templates of Task Plan and Concluding is shown in Appendix \ref{appendix-agent-loop}.

\begin{figure}[t]
\centering
\includegraphics[height=0.46\textwidth]{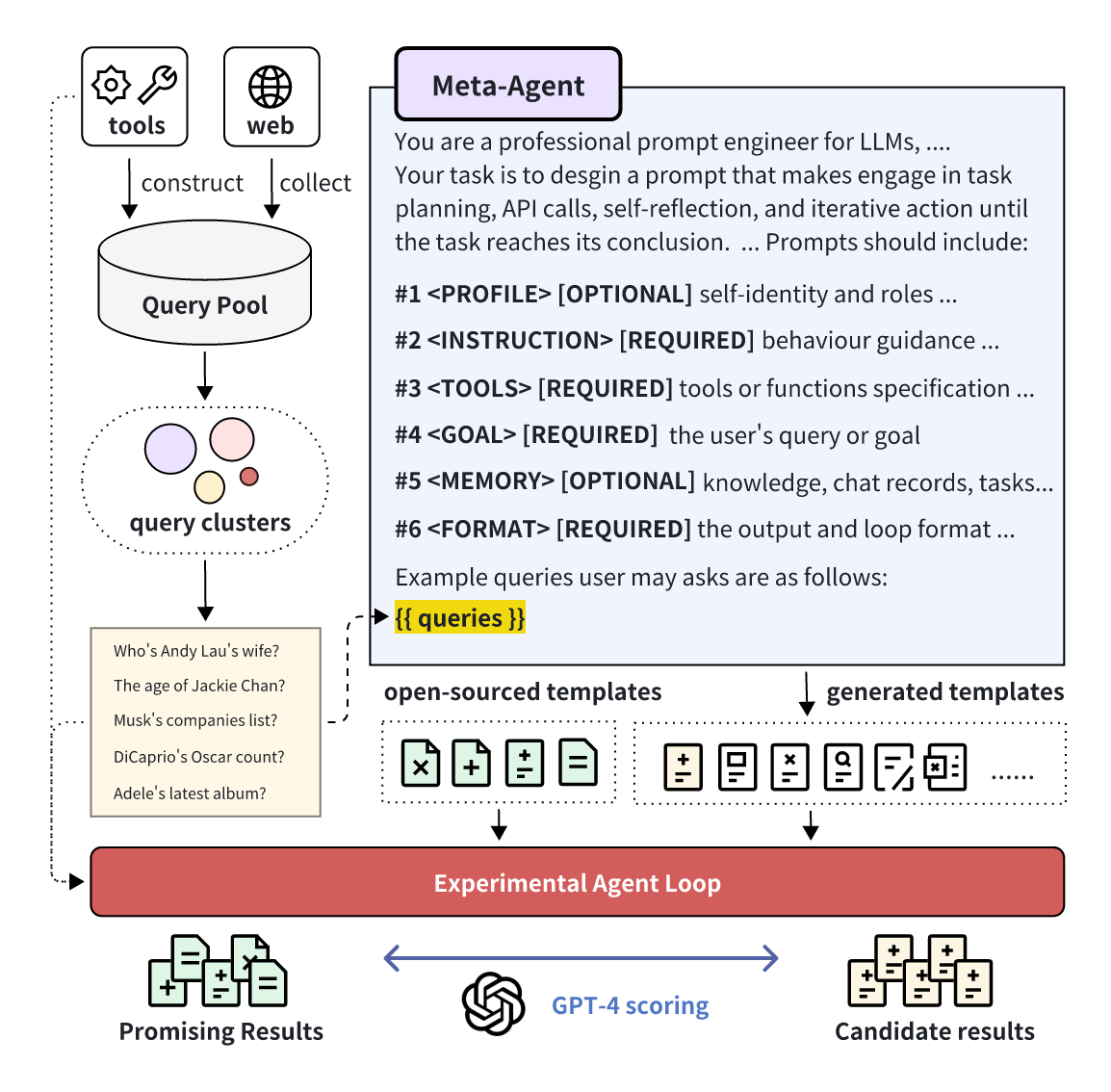}
\caption{The process of how meta-agent crafts agent system templates and validation.}\label{fig:meta-agent}
\end{figure}

\section{Meta-Agent Tuning} \label{sec-mat}

In this section, we introduce the Meta-agent Tuning (MAT) framework, designed to enhance smaller open-source Large Language Models (LLMs) in the 7-13 billion parameter range, with capabilities like planning, reflection, and tool utilization, particularly in information-seeking contexts. The MAT framework comprises three core processes: crafting agent system templates using a meta-agent, validating these templates through an experimental agent loop, and training LLMs with a combination of the generated agent-instruction data and general data.

\subsection{Templates Crafting}
Upon reviewing various open-source agent systems, such as ReAct~\citep{yao-react}, Auto-GPT~\citep{auto-gpt}, ToolLlama~\citep{qin2023toolllm}, and ModelScope-agent~\citep{li2023modelscopeagent}, we observed a shared architectural feature in their system prompts. We distilled this commonality into six components:
(1) \textbf{Profile}, which details the LLM's role, such as "You are a helpful AI Planner".
(2) \textbf{Instructions}, encompassing the constraints and sequence of actions for the agents, for instance, "Iterate no more than five times".
(3) \textbf{Tools}, outlining the format for the tools to be utilized, like a JSON schema with function names, descriptions, and argument details.
(4) \textbf{Memory}, indicating the integration of external knowledge, past tasks, and conversational history; an example being placing the conversation history at the beginning of the prompt.
(5) \textbf{Goal}, specifying the incorporation of user queries or requirements.
(6) \textbf{Format}, illustrating how agents should craft responses and manage iterations, such as returning a JSON object with task names, tool names, and arguments.

To exploit these insights, we designed a meta-agent~\footnote{The full prompt can be found in Appendix~\ref{agent-templates}}, depicted in Figure~\ref{fig:meta-agent}, employing GPT-4 to craft template prompts. Nevertheless, the GPT-4 generated templates were homogenous, lacking in variety. To mitigate this, we introduced seed queries on particular topics to steer GPT-4 towards producing more targeted and diverse templates. Thus, query collection is crucial in this context. We amassed high-quality queries from various sources including ShareGPT~\footnote{https://sharegpt.com/}, prior studies~\citep{qin2023toolllm,li2023modelscopeagent}, and GPT-4 generated queries based on tools following established methodologies~\citep{qin2023toolllm}. After compiling the queries, we segmented them into overlapping clusters, such as celebrity information (as shown in Figure~\ref{fig:meta-agent}). Finally, Meta-agent help us generate an series of system prompt templates for subsequent use.

\subsection{Templates Validation}
\label{sec:Templates Validation}
The primary issue with template generation, as discussed, concerns the quality of the output. Despite the advanced capabilities of GPT-4, some of the resultant templates often exhibit significant flaws, failing to operate in simulated environments or diverging from expected human outcomes. To address this, a validation framework is essential to assess and determine the viability of agent system templates generated by the GPT-4.

Our analysis indicates that open-source agent system templates typically yield higher success rates compared to those generated by GPT-4. By bundling these proven templates, we can enhance the validation process. This process involves a comparative analysis where a subset of queries $Q=\{q_1, q_2, ..., q_n\}$ is applied to both the GPT-generated template $t_g$ and the open-source templates $T_o=\{t_{o,1}, t_{o,2}, ..., t_{o,k}\}$. The performance of these templates is then evaluated in an experimental agent loop powered by GPT-4, producing a set of promising results $R=\{r_{1}^{(i)}, r_{2}^{(i)}, ..., r_{k}^{(i)}\}$ for each query $q_i$. Concurrently, the loop generates a candidate result $r_c^{(i)}$ for $t_g$. We leverage GPT-4's analytical prowess to rate the closeness of $r_c^{(i)}$ to the promising results, represented by a score $\sigma({r_c, r_{j}^{(i)}}) \in \{1, 2, 3, 4, 5\}$.

The template's final score $r_g$ is computed using the formula:
\begin{equation}
score(t_g) = \frac{1}{n}\sum_{i=1}^n \max_{j=1}^k \sigma({r_c^{(i)}, r_{j}^{(i)}})
\end{equation}

To conclude the validation, we establish a threshold $\xi$, approving templates $t_g$ that meet or exceed this benchmark with $score(t_g) \geq \xi$.

\subsection{LLMs Training}
Following the initial phase, we have amassed a collection of high-quality generated agent system templates, open-source templates and templates used in KAgentSys. We proceeded to dispatch a series of queries to each template, employing an experimental agent loop that prompts GPT-4 to produce corresponding prompt-response pairs. This process yielded the agent dataset, denoted as $\mathcal{D}_{agent}$, which, when amalgamated with a diverse instruction dataset $\mathcal{D}_{general}$, facilitates the training of a foundational large language model. We also experiment an appropriate mixture rate, $\eta$, of these two dataset, to apply during this training phase. 

Further details regarding data construction and distribution can be found in Appendix \ref{data-construct}.

\section{Experiments} \label{exp}

This section outlines our experimental approach, detailing our specially curated instruction-tuning dataset and introducing the KAgentBench. We present the performance of different large language models (LLMs) on our benchmark, illustrating models' agent capabilities across various perspective. We also collect some factual or time-aware queries to show how different models perform on different systems.

\subsection{Dataset}
\begin{figure}[t]
\centering
\includegraphics[height=0.45\textwidth]{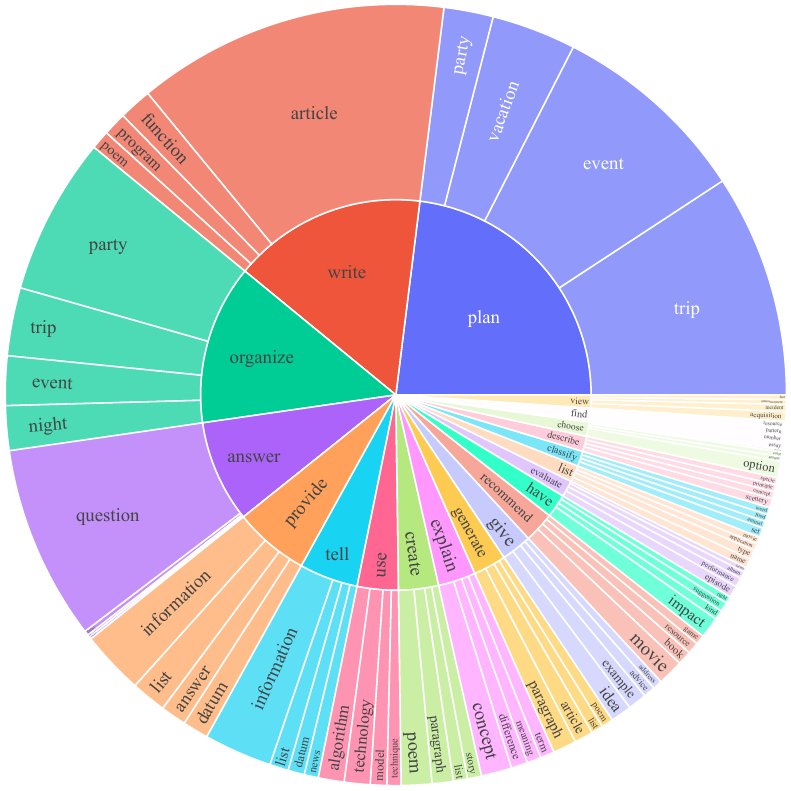}
\caption{The visualization of the query diversity in our instruction tuning data. The inner circle represents the root verb of questions, and the outer circle lists the direct noun objects
of the questions.}\label{fig:query-dist}
\end{figure}

Our instruction-tuning dataset was compiled by aggregating and generating tools and queries from prior research~\citep{qin2023toolllm,li2023modelscopeagent,auto-gpt}. This was augmented with additional queries focused on factual content. Following the template crafting phase, we sampled approximately 50 queries for each generation, which yielded a substantial number of candidate templates. These were then vetted through an experimental agent loop, with those achieving a threshold score $\xi$ being selected for use~\footnote{See appendix ~\ref{agent-templates} for more scoring details.}. We incorporate open-source templates, the KAgentSys template, and our generated templates alongside the sampled queries into the experimental agent loop. This resulted in a collection of prompt-response pairs, comprising 120,917 queries and 18,005 templates, as detailed in Table \ref{table:train-data}. Additionally, to avoid overemphasis on specialized capabilities of LLMs, we collated 43,099 queries from diverse domains such as open-domain question answering, chit chatting, role-playing, mathematics, etc. from different sources. The distribution of these queries is depicted in Figure \ref{fig:query-dist}.

The KAgentBench was constructed through a rigorous selection of queries, tools, templates, and memory elements, with a detailed breakdown presented in Figure \ref{table:kagentbench}. The benchmark includes 63.04\% of queries in Chinese, with the remainder in English.
Each query is paired with five system prompt templates, with the exception of the ``Profile'' type, which increases the instance count. Four-to-five ground truth responses generated by different templates are associated with each instance, predicated on the assumption that these truths are independent of the prompt templates. The benchmark features 614 unique tools in toal, and 9.26 tailored in average to individual queries. Memory is categorized into four types: none, conversational turns, task history, or external knowledge. Among them, no memory accounts for 20.64\%, conversational turns accounts for 48.12\%, task history accounts for 16.23\%, and external knowledge accounts for 15.01\%. Particularly, in memory, conversational turns, task history, and external knowledge data all contain information types that are related, unrelated, and conflicting with the query. ``Reflection'' is distinguished from ``Planning \& Tool-use'' by the need to address incorrect information or discrepancies in previous tasks or external knowledge, requiring LLMs to reassess current information for more strategic planning, such as formulating alternative search queries. 
The ``Profile'' dimension involves crafting instructional conversations within specific roles, for which GPT-4 generates variations under different settings. Post-creation, human annotators refined each ground truth to ensure a high-quality benchmark.

Furthermore, we amassed 200 factual and time-aware queries in Chinese to serve as initial inputs for an agent system, with the option to employ human annotators to assess the accuracy and the quality of the system's final responses.

\begin{table}[!t] 
 \centering
 \begin{tabular}{lcccc} 
  \toprule
  Type & \#Instance & \#Queries & \#Templates & Avg. \#Steps\\
  \midrule
  Agent & 224,137 & 120,917 & 18,005 & 1.85 \\
  General & 43,099 & 43,099 & - & - \\
  \bottomrule 
 
 \end{tabular} 
 \caption{
    The statistics of the mixture of our instructional data. 
 } 
\label{table:train-data}
\end{table}

\begin{table*}[!t] 
 \centering
 \begin{tabular}{l|ccccccccc} 
  \toprule
  type & \#Queries & \#Inst & Avg. \#Ground & Avg. \#Tools & Avg. \#Turns & Avg. \#Tasks & Avg. Len-Know  & Metric \\
  \midrule

  Planning \&  Tool-use & 320 & 1,317 & 4.12 & 8.68 & 1.51 & 2.21 & 245.31 & ROUGE-L, EM \\
  Reflection & 68 & 272 & 4 & 12 & 1 & 3.97 & 1369.04 & ROUGE-L, EM \\
  Concluding & 245 &  1,225 & 5 & - & 1.52 & 2.14 & 923.96 & ROUGE-L \\
  Profile & 433 & 433 & 5 & - & 1.99 & - & - & ROUGE-L \\
  
  \bottomrule 
  
 \end{tabular} 
 \caption{
    Overall statistics of KAgentBench. ``\#Inst'' indicates the count of prompts provided for LLMs to generate responses. ``\#Ground'' denotes the quantity of ground truth responses corresponding to each prompt. ``\#Tools'' refers to the quantity of tools utilized during the planning phase. ``\#Turns'' represents the total conversation history turns retained in memory plus the current turn, while ``\#Task'' specifies the count of historical task-related steps maintained plus the current one and ``Len-Know'' quantifies the token length of external knowledge incorporated within the memory.
 } 
\label{table:kagentbench}
\end{table*}

\subsection{Benchmark Evaluation}

In the planning stage, the model is tasked with producing three key outputs: 
(1) \textbf{Thought} $T_h = \{t_1, t_2, ..., t_3\}$ is the text generated by LLMs to decide what it will do next, e.g. \texttt{"I will search Andy Lau's personal information"},
(2) \textbf{Tool-name} $T_n$ is the designated tool for the task, e.g. \texttt{web\_search}
(3) \textbf{Tool-args} $T_a = \{ a_1, v_1, a_2, v_2, ...\}$ is the list of (argument\_name, arg\_value)s. e.g. \texttt{[keywords, Andy Lau, topk, 5]}.
To assess different agent capabilities, we introduce specific datasets and metrics:

\noindent\textbf{Planning} ability evaluation involves adhering to the ``Unity of knowledge and action'' principle~\citep{ymwangunity}. It requires considering both the thought and the tool-name:
\begin{equation}
    S_{plan} = \frac{1}{M}\sum_{j=1}^{M}\max_{i=1}^N EM(T_{n, i}, T'_{n, j}) \cdot \mathcal{T}(T_{h, i}, T'_{h, j}) 
\end{equation}
Here $T_{*, i}$ is the $i$-th ground truth result. $T'_{*, j}$ is predicted result used by $j$-th prompt templates. $M$ denotes the number of templates and $N$ is the number of ground truths. $\mathcal{T}$ is any metrics used for text generation evaluation, e.g. ROUGE score.

\noindent\textbf{Tool-use} proficiency is gauged by both the correct selection (tool-name) and application of tools (tool-args): 
\begin{equation}
    S_{tool-use} = \frac{1}{M}\sum_{j=1}^{M}\max_{i=1}^N EM(T_{n, k}, T'_{n, j}) \cdot \sum_{k=1}^{K_i} EM(a_{k, i}, a_{k, j}')\cdot\mathcal{T}(v_{k, i}, v_{k, j}')
\end{equation}

\noindent\textbf{Reflection} draws on a distinct dataset and combines the scoring methods for planning and tool-use, adjusted by a penalty factor $\mathbf{p}\in\{0, 1\}$. A penalty applies if the generated tool-names and arguments repeat from a previous task; 
\begin{equation}
    S_{reflection} = (1 - \mathbf{p}) \cdot (0.3 \cdot S_{plan} + 0.7 \cdot S_{tool-use}) 
\end{equation}

\noindent \textbf{Concluding} stage involves generating a conclusion $T_c$, simply a word sequence, hence the scoring simplifies to:
\begin{equation}
    S_{concluding} = \frac{1}{M}\sum_{j=1}^{M}\max_{i=1}^N \mathcal{T}(T_{h, i}, T'_{h, j})
\end{equation}

\noindent \textbf{Profile} benchmark utilizes a distinct dataset for role-based instructional conversations. The response ($T_r$) also constitutes a word sequence, but with a singular template and $N$ ground truths, leading to this scoring formula:
\begin{equation}
    S_{profile} = \max_{i=1}^N \mathcal{T}(T_{h, i}, T'_{h, j})
\end{equation}

The aggregate score is a weighted sum of the five metrics, with careful consideration given to the weights:

\begin{align}
    S_{overall} = 0.25 \cdot S_{planning} + 0.35 \cdot S_{tool-use} + 0.1 \cdot S_{reflection} \nonumber \\
    + 0.2 \cdot S_{concluding} + 0.1 \cdot S_{profile} 
\end{align}

The performance of various LLMs is shown in Table~\ref{table:results-devset}. Among all aspects, GPT-3.5 surpasses all experimented open-source models. After meta-agent tuning, Qwen-7B and Baichuan2-13B exhibit significant improvements of 14.22 and 19.71, respectively, exceeding GPT-3.5's performance. Llama2, ToolLlama, and AgentLM obtain poor results in planning, tool-use, and reflection due to their inability to generate the correct formatting for each template and their limited capacity to handle Chinese contexts. ToolLlama's performance is worse in planning compared to tool-use, as we discovered that their training data consistently has a ``thought'' field set to none, affecting the planning metric computation. Therefore, even with clear prompt instructions, models tuned to specific prompt templates consistently fail. Furthermore, we observed that reflection is more challenging than planning and tool-use since it necessitates higher model intelligence and longer task history, along with an expanded knowledge length.

% \textcolor{red}{Need analysis}
% \textcolor{red}{the open-sourced models cannot outperform closed-source such as GPT 3.5; explain why llama2 get poor results on planning, tool-use, reflection; explain why toollama get extremly poor results on planning and reflection; explain why agentLM get poor results(tuned by specific six tasks with the same prompts, lost generalization ability to other agent system prompts); explain Qwen-7B perform the best (they put) react data balabala. After mat the performance get huge improvements}

\begin{table*}[!t] 
 \centering
 \begin{tabular}{l|c|cccccc} 
  \toprule
  ~ & Scale & Planning & Tool-use & Reflection  & Concluding & Profile & Overall Score \\
  \midrule
  GPT-3.5-turbo & - & 18.55 & 26.26 & 8.06 & 37.26  & 35.42 & 25.63 \\
  % GPT-4 & ~ & ~ & ~ & ~ & ~ & ~ \\
  Llama2 & 13B & 0.15 & 0.44 & 0.14 & 16.60 & 17.73 & 5.30 \\

  ChatGLM3 & 6B & 7.87 & 11.84 & 7.52 & 30.01 & 30.14 & 15.88 \\
  % Qwen & 14B & 10.56 & 11.47 & OnInf & 46.99 & 30.06 & ~ \\
  Qwen & 7B & 13.34 & 18.00 & 7.91 & 36.24 & 34.99 & 21.17 \\
  Baichuan2 & 13B & 6.70 & 16.10 & 6.76 & 24.97 & 19.08 & 14.89 \\
  ToolLlama& 7B & 0.20 & 4.83 & 1.06 & 15.62 & 10.66 & 6.04 \\
  AgentLM & 13B & 0.17 & 0.15 & 0.05 & 16.30 & 15.22 & 4.88 \\
  \midrule
  Qwen-MAT & 7B & 31.64 & 43.30 & 33.34 &  44.85  &  44.78  & 39.85 \\
  % Qwen-MAT & 14B & 29.91 & 28.70 & OnInf & 53.76 & 40.44 & ~ \\
  % Baichuan2-MAT & 7B & OnInf & OnInf & OnInf & ~ & ~ & ~ \\
  Baichuan2-MAT & 13B & 37.27 & 52.97 & 37.00 & 48.01 & 41.83 & 45.34 \\
 \bottomrule 
 
 \end{tabular} 
 \caption{
    Experimental results of different LLMs on KAgentBench.
 } 
\label{table:results-devset}
\end{table*}

\begin{table*}[!t] 
 \centering
 \begin{tabular}{l|c|cccc} 
  \toprule
  ~ & Scale & NoAgent & ReACT  & Auto-GPT & KAgentSys \\
  \midrule
  GPT-4 & - & 57.21\% (3.42) & 68.66\% (3.88) & 79.60\% (4.27)  & 83.58\% (4.47)  \\
  GPT-3.5-turbo & - & 47.26\% (3.08) & 54.23\% (3.33) & 61.74\% (3.53) & 64.18\% (3.69) \\
  Qwen & 7B & 52.74\% (3.23) & 51.74\% (3.20) &  50.25\% (3.11) & 54.23\% (3.27)  \\
  % Qwen & 14B & 61.69\% (3.54) & 59.20\% (3.48) &  50.25\% (3.11) & 63.68\% (3.72)  \\
  Baichuan2 & 13B & 54.23\% (3.31) &  55.72\% (3.36) & 57.21\% (3.37) & 58.71\% (3.54)  \\
  \midrule
  Qwen-MAT & 7B & - & 58.71\% (3.53) & 65.67\% (3.77) & 67.66\% (3.87) \\
  % Qwen-MAT (14B) & -  & OnEval & PreEval & PreEval \\
  Baichuan2-MAT & 13B &  - & 61.19\% (3.60) & 66.67\% (3.86) & 74.13\% (4.11) \\
  \bottomrule 
 
 \end{tabular} 
 \caption{
    Human evaluation results for different agent systems of information seeking queries. Each result cell shows the pass rate (\%) and the average score (in parentheses).
 } 
\label{table:human-eval}
\end{table*}

\subsection{Human Evaluation}
As previously mentioned, we collect approximately 200 queries as the initial input for each agent system equipped with various LLMs as the brain. Each system ultimately outputs a conclusion in response to the input. Human annotators are employed to assess the truthfulness and quality of the responses. The scoring consists of five levels, defined as follows: 1 point - significant inaccuracies in the core information, with most or all being incorrect; 2 points - core information is incorrect, or over 60\% of the content contains errors; 3 points - core information is accurate, but 10\% to 60\% of the supplementary information is incorrect; 4 points - minor discrepancies, with the core information being accurate and errors in the supplementary information limited to 10\% or less; 5 points - flawless. A response with a score of 4 or 5 points is considered to have passed.

Owing to resource limitations, we only evaluate the two most popular open-source agent systems, ReAct and AutoGPT, in comparison to KAgentSys. The results are displayed in Table \ref{table:human-eval}. Generally, we observe that agent systems yield better results than directly querying LLMs. Qwen-7B and Baichuan2-13B perform well without an agent, possibly due to the integration of Chinese factual data during the pretraining or supervised fine-tuning stage. AutoGPT outperforms ReAct, as the prompt templates are more complex, and the JSON output format is more stable. Our KAgentSys achieves the best results, regardless of which backend LLM is used. There is a considerable gap between GPT-4 and other models, with the open-sourced models performing worse. However, following meta-agent tuning, the open-source models show significant improvements; Qwen-7B has an 11.9\% increase in average pass rate across three systems and a 0.53 increase in average score, while Baichuan-13B sees a 10.1\% increase in pass rate and a 0.43 increase in average score. After meta-agent tuning, the best open-source model achieves 74.13\%, which is 10\% higher than GPT-3.5 and close to GPT-4's 83.58\%.

\subsection{Ablation Study}
To further demonstrate the effectiveness of KAgentSys, we modify several key components: 1) we replace the memory bank with a simple memory mechanism where tasks, conversations, and external knowledge are inserted into the prompt by truncation; 2) we substitute the hybrid search-browse toolset with the basic \texttt{web\_search} and \texttt{web\_browse} modules used in Auto-GPT. This simplified version is called KAgentSys-lite, which will also be open-sourced. We evaluate KAgentSys-lite using the aforementioned human evaluation method and employ Baichuan2-MAT as the cognition core. The results reveal a 5.97\% drop (74.13\% → 68.16\%) in pass rate and a 0.25 decrease (4.11 → 3.86) in average score. Moreover, to assess the generalizability of our Meta-agent Tuning approach on unseen templates, we exclude all open-source agent templates from the training set. Following training, the performance of Baichuan2-MAT in KAgentBench shows moderate decreases in overall score (45.34 → 42.12, 3.22\% drop) metrics. These results indicate that open-source agent templates is not a key component of our training process and demonstrate the models after Meta-agent Tuning are robust on unseen templates. To evaluate KAgentLMs's generalization ability for unseen tools , we conduct evaluations on the KAgentBench subset, focusing on functions not present in the training set. The results reveal that, in terms of the plan metric, the Baichuan2-MAT model shows a 15.71-point improvement over GPT-3.5-turbo (21.34 → 37.05). Similarly, for tool-use metric, Baichuan2-MAT achieves a 14.04-point enhancement compared to GPT-3.5-turbo (35.56 → 49.60). These findings demonstrate the model's robust generalization ability for unknown tools.

\begin{figure*}[t]
\centering
\includegraphics[height=0.7\textwidth]{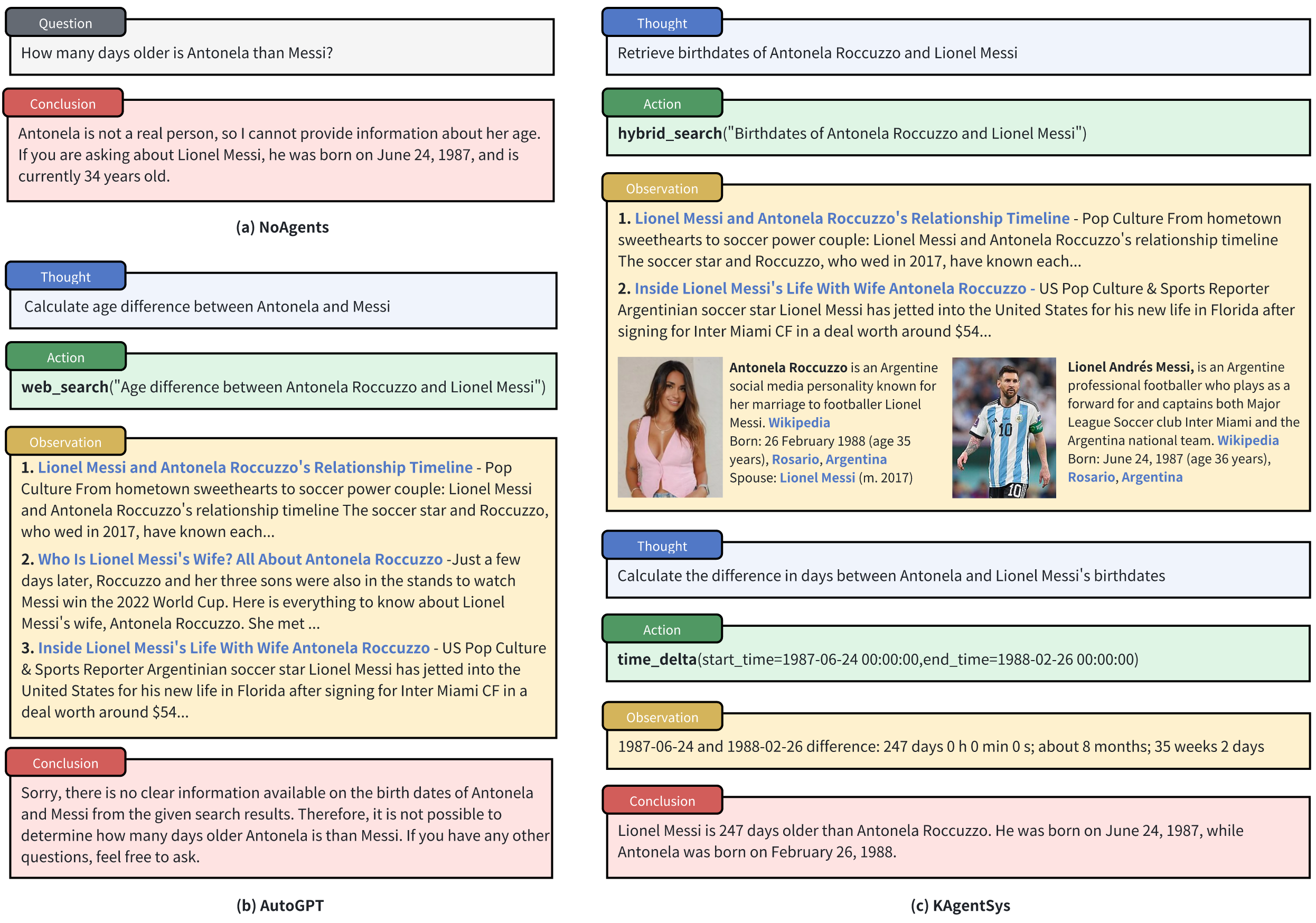}
\caption{The decision path of different systems when answer the question ``how many days older is Antonela than Messi''.}\label{fig:sys-case}
\end{figure*}

\subsection{Case Study}

We have conducted case studies to evaluate the performance of our proposed system and modeling approach. During these studies, we observed that GPT-3.5 tends to generate repetitive planning stages and employ the same strategies, even when no new information is gleaned or an answer has already been found. Contrarily, our modified approach demonstrates enhanced reflective capabilities and an ability to discontinue redundant processes effectively. Additional details can be found in Appendix ~\citep{appendix-model-comparison}, referenced in Figure ~\ref{fig:model_case}.

Regarding the system analysis, traditional search engines exhibit limitations when faced with long-tail or trending questions. For instance, as depicted in Figure ~\ref{fig:sys-case}, when querying the age difference between Antonela and Messi, two issues arise: (1) the trendiness of ``Messi and his wife'' skews search results towards news articles that tempt user engagement with irrelevant content, such as relationship timelines; and (2) the question, which pertains to their respective birthdates (a detail not widely sought), is a ``long-tail'' search query. Large language models (LLMs) alone struggle with this, as they might recall Messi's birthdate but forget Antonela's. Combining LLMs with a search engine also falls short, as the surfaced information, while related, fails to precisely address the inquiry.
Our approach overcomes these hurdles by incorporating entity linking and extracting relevant details from resources like Wikipedia. Specifically, our system first retrieves Messi and his wife's birthdates, then accurately computes the time difference using the \texttt{time\_delta} tool, delivering the correct response to the posed question.
 \section{Related Work} \label{review}

\subsection{Large Language Models}

The field of Large Language Models (LLMs) has recently undergone a significant revolution, showcasing their prowess across a variety of downstream domains and tasks~\cite{tom-gpt3,aakanksha-palm,touvron2023llama}. 
Enhanced capabilities are achieved through advanced prompting ~\cite{jason-cot} and instruction fine-tuning~\cite{ouyang-instruct-tuning,tay-flan}.
To make LLMs accessible for anyone, numerous open-source projects have been launched, including Llama~\cite{touvron2023llama,touvron2023llama2}, Alpaca~\citep{stanford-alpaca}, Vicuna~\citep{vicuna2023}, etc. Many LLMs are optimized for Chinese language processing, such as such as ChatGLM~\citep{zeng2023glm-130b}, Qwen~\citep{jinze-qwen}, Baichuan~\citep{baichuan2023baichuan2}. 
These models, generally ranging from 6B to 14B parameters, also achieve promising results on many benchmarks. Our study investigates their effectiveness within agent systems, in comparison to closed-source alternatives such as GPT-4.

\subsection{LLMs Powered Agents}
Recent research has increasingly shown that LLMs can be instrumental in developing AI agents~\citep{lilianweng-autoagents,norman-unified-agent,xi2023rise,wang2023survey}. Examples such as ToolFormer~\citep{timo-toolformer}, HuggingGPT~\citep{shen-hugginggpt}, Gorilla~\citep{patil2023gorilla}, and ToolLlama~\citep{qin2023toolllm} demonstrate LLMs' proficiency in tool utilization. Projects like Auto-GPT~\citep{auto-gpt}, BabyAGI~\citep{baby-agi}, and ModelScope-agent~\citep{li2023modelscopeagent} highlight the autonomous capabilities of LLM-powered agents in fulfilling human tasks. Generative Agents~\citep{generative-agents} are known for mimicking human interactions within immersive environments. Voyager~\citep{wang2023voyager} showcases these agents' ability to acquire diverse skills and engage in continuous exploration in settings like Minecraft. Additionally, AutoGen~\citep{wu2023autogen} and ChatDev~\citep{qian2023communicative} facilitate agent-to-agent communication, aiding in application development. Our study focuses on constructing an agent system adept at using external search engines or encyclopedic resources for completing information-seeking tasks.
 
\subsection{Information seeking systems}
Our research contributes to the domain of conversational information-seeking systems. \citet{radlinski2017theoretical} articulated the core attributes of these systems, while \citet{thomas2017misc} introduced the Microsoft Information-Seeking Conversation (MISC) dataset, which presents dialogues where humans serve as intermediaries in information-seeking scenarios. Agents utilizing reinforcement learning have been effectively employed in this area as well \citep{pan-rl-prf}. Of particular interest is the emergence of Large Language Model (LLM)-driven methods and systems, such as WebGPT \citep{webgpt}, AutoGPT \citep{auto-gpt}, and innovative approaches to web navigation \citep{yao2022webshop,deng2023mind2web,izzeddin23webagent}. Our study harnesses the capabilities of LLMs to construct an agent system that automates the gathering of pertinent information, thereby satisfying the informational needs of users.

\section{Conclusion} \label{conclusion}

In this paper, we present \systemname, composed of three integral components. The agent system KAgentSys employs a planning-concluding procedure, leveraging LLMs as the core cognitive unit, a memory bank, and a hybrid time-aware search-browse toolkit to fulfill user queries effectively. To investigate the potential of open-source, smaller LLMs in exhibiting agent capabilities such as planning, reflection, and tool utilization, we propose a Meta-agent Tuning (MAT) strategy and produce a suit of LLMs called KAgentLMs. Furthermore, we establish a comprehensive benchmark KAgentBench to assess the aforementioned capabilities of LLMs and gather an additional \~200 queries to further appraise system-level performance. Our extensive experiments demonstrate that our system outperforms other open-source agent systems. And after MAT, even LLMs in the range of 6B-13B exhibit results comparable to GPT-4.

\bibliographystyle{ACM-Reference-Format}
\bibliography{sample-base}
\newpage
\appendix

\section{KAgentSys}

\subsection{Memory Bank} \label{appendix-memory-bank}

In this chapter, we will detail the implementation mechanism of the memory module, which plays a crucial role in the system, storing user context information during the dialogue process of an intelligent agent. The memory module is divided into conversation memory, task memory, and knowledge memory, based on the type of information stored. These components have been mentioned in the main body of the paper, and now we will further elaborate on the specific implementation of the memory bank mechanism.

The foundation of the memory bank mechanism is a vector database, serving as the underlying infrastructure to support the storage of three different types of information. Information from conversation memory, task memory, and knowledge memory undergoes slicing and vectorization before being stored in this shared vector database, facilitating subsequent retrieval and updates.

In our approach to data vectorization and retrieval, we have integrated the techniques of Embedding vectorization and Elasticsearch (ES). The embedding method transforms document segments into dense embedding vectors, capturing their semantic essence. In contrast, Elasticsearch represents these segments as sparse vectors through an inverted index, focusing on keyword identification. For retrieval, the embedding-based method conducts searches by measuring semantic similarities, while Elasticsearch performs retrieval centered on keyword matching. Our current strategy merges these two methods by concatenating the results from both the embedding and Elasticsearch retrievals. This process involves merging, deduplicating, and then seamlessly integrating the outcomes from both retrieval systems to enhance the overall effectiveness. By integrating these two vectorization techniques, the memory bank can achieve more comprehensive and accurate information recall.

Building on these infrastructures, we have developed specific retrieval and update mechanisms for the three types of memory. During the dialogue process of an intelligent agent, each type of memory has its unique retrieval mechanism. For knowledge memory, we increase the number of recall results to access more relevant information. For conversation memory, we ensure the recall of contextually relevant dialogue turns related to the query. For task memory, we focus on recalling information related to the most recent tasks, adding a temporal weight. These unique retrieval strategies for each type of memory are designed to optimize the retrieval efficiency and accuracy of specific types of information.

\begin{figure*}[t]
\centering
\includegraphics[width=\linewidth]{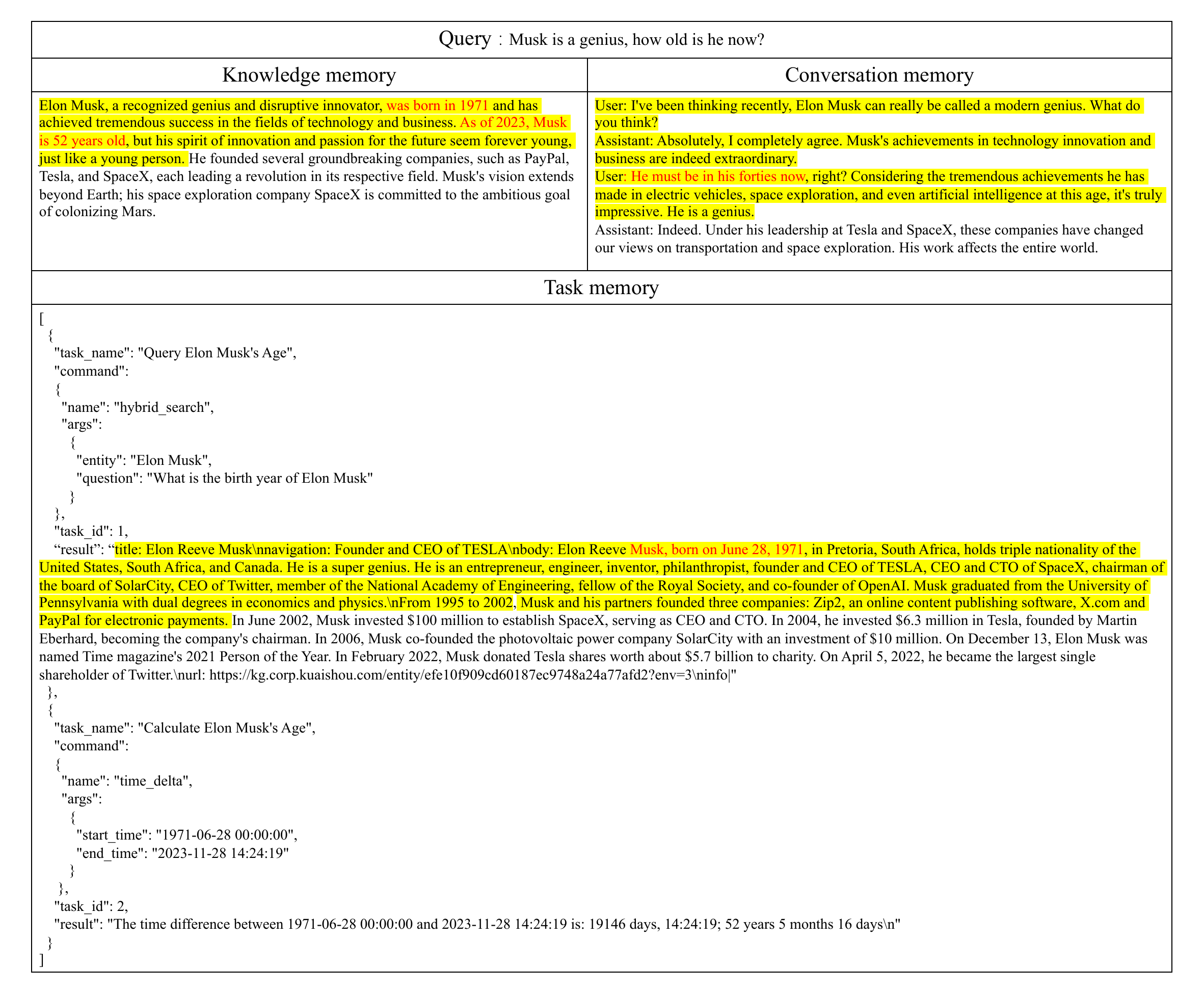}
\caption{Different types of memory retrieved after a query "Musk is a genius, how old is he now?" search. The yellow highlighted section represents the text retrieved based on the query, and the red text signifies information pertinent to the text and query response.}\label{fig:memory bank}
\end{figure*}

For example, for the query "How old is Elon Musk," the information retrieved from different memory types in the memory bank is shown in Figure \ref{fig:memory bank}. The background color marked in yellow represents parts related to the query during the retrieved, while the text marked in red indicates information related to the text and the query answer.

Conversation memory and task memory are continuously updated as the dialogue progresses, ensuring that the intelligent agent can effectively handle dynamically changing dialogue situations and task requirements. This design not only enhances the flexibility of the memory module but also improves its application efficiency and accuracy in actual dialogue scenarios.

Finally, three types of memory data are concatenated directly to form a comprehensive memory information set. This consolidated memory information is then seamlessly integrated into the designated memory slots within the prompt.

\subsection{Tool Library} \label{appendix-tool-library}

We follow openai’s function format~\footnote{https://platform.openai.com/docs/guides/function-calling} to define our tools as follows:
\begin{lstlisting}[language=json,firstnumber=1]
{
  "name": "hybrid_search",
  "description": "Perform hybrid search including 1) web_search 2) wiki_search 3) video_search.",
  "parameters": {
    "type": "object",
    "properties": {
      "text": {
        "type": "str",
        "description": "Search query."
      }
    }
  },
  "required": [
    "text"
  ]
}

{
  "name": "browse_website",
  "description": "Browse a specific website using the provided URL link. ",
  "parameters": {
    "type": "object",
    "properties": {
      "url": {
        "type": "str",
        "description": "The website's URL link."
      },
      "question": {
        "type": "str",
        "description": "The specific content or topic sought on the website."
      }
    }
  },
  "required": [
    "url",
    "question"
  ]
}

{
  "name": "browse_wiki",
  "description": "Browse the content of a specified Wikipedia page URL.",
  "parameters": {
    "type": "object",
    "properties": {
      "url": {
        "type": "str",
        "description": "The Wikipedia page URL."
      },
      "question": {
        "type": "str",
        "description": "The information of interest on the wikipage."
      }
    }
  },
  "required": [
    "url",
    "question"
  ]
}


{
  "name": "browse_aspect",
  "description": "Browse the video list given an entity id and aspect of kuaipedia by previous search.",
  "parameters": {
    "type": "object",
    "properties": {
      "entity_id": {
        "type": "str",
        "description": "The ID of entity"
      },
      "aspect": {
        "type": "str",
        "description": "The aspect of the entity to search"
      }
    }
  },
  "required": [
    "entity_id",
    "aspect"
  ]
}



{
  "name": "browse_video",
  "description": "Browse the video to get detailed information",
  "parameters": {
    "type": "object",
    "properties": {
      "video_id": {
        "type": "str",
        "description": "The ID of the video"
      }
    }
  },
  "required": [
    "video_id"
  ]
}

{
  "name": "get_weather_info",
  "description": "Retrieve weather information for specified locations and dates.",
  "parameters": {
    "type": "object",
    "properties": {
      "location": {
        "type": "str",
        "description": "Locations in English separated by commas, e.g., \"Beijing,Vancouver,...,Chicago\"."
      },
      "start_date": {
        "type": "str",
        "description": "Start date in format \"yyyy-MM-dd\"."
      },
      "end_date": {
        "type": "str",
        "description": "End date in format \"yyyy-MM-dd\"."
      },
      "is_current": {
        "type": "str",
        "description": "\"yes\" or \"no\" indicating if current time's weather is desired."
      }
    }
  },
  "required": [
    "location",
    "start_date",
    "end_date",
    "is_current"
  ]
}

{
  "name": "get_calendar_info",
  "description": "Retrieve calendar details between specified dates.",
  "parameters": {
    "type": "object",
    "properties": {
      "start_date": {
        "type": "str",
        "description": "Start date in the format \"yyyy-MM-dd\"."
      },
      "end_date": {
        "type": "str",
        "description": "End date in the format \"yyyy-MM-dd\"."
      }
    }
  },
  "required": [
    "start_date",
    "end_date"
  ]
}

{
  "name": "time_delta",
  "description": "Calculate the time interval between two timestamps.",
  "parameters": {
    "type": "object",
    "properties": {
      "start_time": {
        "type": "str",
        "description": "format of \"yyyy-MM-dd HH:mm:ss\"."
      },
      "end_time": {
        "type": "str",
        "description": "format of \"yyyy-MM-dd HH:mm:ss\"."
      }
    }
  },
  "required": [
    "start_time",
    "end_time"
  ]
}

{
  "name": "get_solar_terms_info",
  "description": "Retrieve solar terms in Chinese for a given year. ",
  "parameters": {
    "type": "object",
    "properties": {
      "year": {
        "type": "int",
        "description": "Target year for query."
      }
    }
  },
  "required": [
    "year"
  ]
}

{
  "name": "get_holidays_info",
  "description": "Retrieve dates and arrangements for all legal holidays in China for a given year. ",
  "parameters": {
    "type": "object",
    "properties": {
      "year": {
        "type": "int",
        "description": "Target year for query."
      }
    }
  },
  "required": [
    "year"
  ]
}

\end{lstlisting}

\subsection{Agent Loop} \label{appendix-agent-loop}
As mentioned in the paper, we use different prompt templates for the stage of planning and the stage of concluding. All of the tools in the toolset will be put under ``Command'' section, plus one ``task\_complete'' command. The maximum iteration number, memory block and user's goal or query will be put into the placeholders in these two templates.

\begin{lstlisting}[language=json,firstnumber=1]
##  KAgentSys Planning prompt

You are an AI assistant, and your role is to help humans solve their problems. Currently, you are in the task planning phase, where you will be given specific goals or problems to address. Your decisions will be executed independently without relying on human assistance. Please utilize LLM's advantages and pursue efficient strategies for task planning.
1. You have a short-term memory of approximately 4,000 characters.
2. You do not require assistance from users.
3. You can use the reference tools mentioned when planning.
4. You have the abilities to perform internet searches, aggregate information, and discern between genuine and fake information.
5. Remain humble and, if unsure about an issue, make use of commands when possible but minimize their usage and avoid repetition.
6. When drawing conclusions from your knowledge or historical memory, be clever and efficient in task completion and conclusion.
7. Regularly engage in constructive self-criticism to reflect on past decisions and strategies and improve your approach.
8. You can think and plan up to {{ max_iter_num }} steps, so strive to plan tasks as efficiently as possible.
9. You have the capability for reflection; if a completed task and its results cannot provide the necessary information to answer a question or achieve a goal, continue planning but avoid repeating previous tasks.
10. If information from Knowledge Info is available, prioritize using it to answer questions. If the content of Knowledge Info cannot resolve the issue, only then may you use the tools.

Commands:
1:{"name": "web_search", "description": "Perform an internet search.", "parameters": {"type": "object", "properties": {"text": {"type": "str", "description": "Search query."}}}, "returns": {"description": "Multiple webpage links along with brief descriptions.", "type": "str"}, "required": ["text"]}
...

Current time: {{datetime.now()}}
Conversation History: {{memory.conv_history}} 
Complete tasks: {{memory.complete_tasks}}
Knowledge Info: {{memory.know_info}}
Goal: {{goal}}
Based on the goal and existing tasks, plan a new Task (no repetitions), and you can only generate the Task in the following json list format:
{"task_name": "task description",  "command": {"name": "command name", "args": {"arg name": "value"}}}
Ensure that the Task can be parsed by Python's json.loads function. If the already completed Tasks are sufficient to answer the goal, then try to generate the Task to complete it as much as possible. Otherwise, create another Task. A new Task:

\end{lstlisting}

\begin{lstlisting}[language=json,firstnumber=1]
##  KAgentSys Concluding prompt

You are an AI assistant, and you can help humans solve their problems.
The current stage is the concluding stage. In the previous interactions, you have already found some information by searching on your own for the user's given goals and problems. You need to integrate this information and provide the final conclusion in Chinese.
If there is information from Knowledge info, and the information can answer the question, you can use the Knowledge info information as much as possible to answer the question without using external tool results or creating your own content.
1. The information you search for comes from many sources and may be redundant.
2. When the information obtained from different tools conflicts, you should follow a certain priority (Knowledge info > Wiki > search) to resolve the conflict.

Current time: {{datetime.now()}}
Conversation History: {{memory.conv_history}} 
Complete tasks: {{memory.complete_tasks}} 
Knowledge Info: {{memory.know_info}}

Goal: {{goal}}
Generate helpful answers for users:
\end{lstlisting}
% \begin{figure*}[t]
% \centering
% \includegraphics[height=1.3\textwidth]{images/kagentsys-prompts.pdf}
% \caption{The prompt templates of KAgentSys.}\label{fig:kagentsys-prompts}
% \end{figure*}

\section{Data Construction} \label{data-construct}

\subsection{Tools and Queries}
For Tools, beyond the pre-defined tools in the KAgentSys tool library, we integrated high-quality and open-source tools, specifically those from ReACT~\citep{yao-react} and AutoGPT's~\citep{auto-gpt} public code, as well as tools from the public datasets of Toolllama~\citep{qin2023toolllm} and Modelscope~\citep{li2023modelscopeagent}. Additionally, following Toolllama's methodology, we have also constructed some tools using the self-instruct approach. As for queries, we collected a substantial number of high-quality queries from open-source communities, including ShareGPT, Alpaca~\footnote{https://crfm.stanford.edu/2023/03/13/alpaca.html}, Toolllama~\citep{qin2023toolllm}, Modelscope~\citep{li2023modelscopeagent}, among others. 

\begin{figure}[t]
\centering
\includegraphics[width=1\columnwidth]{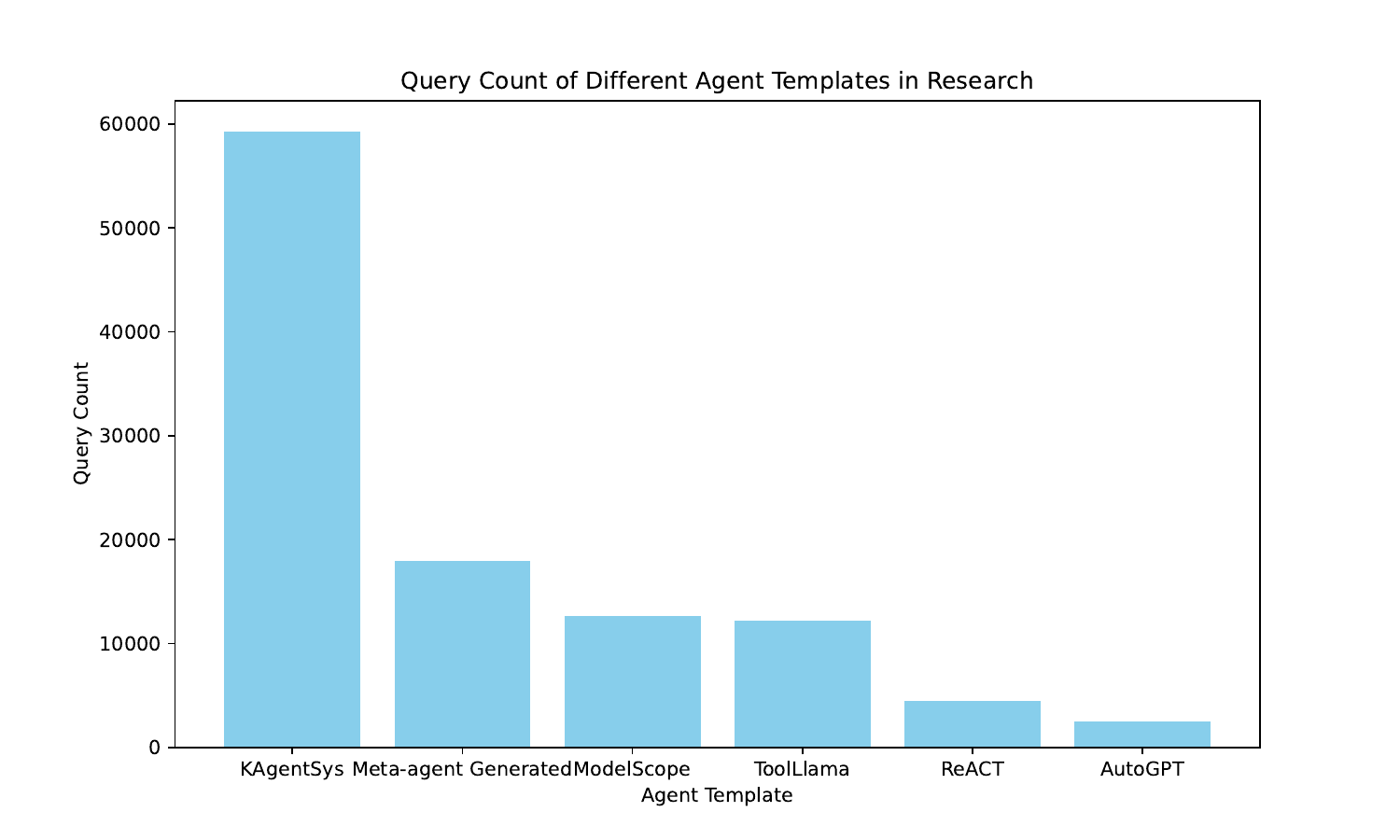}
\caption{Distribution of queries under various agent templates in the training set.}\label{fig:agent distribution}
\end{figure}

\subsection{Agent Templates} \label{agent-templates}
  In our study, we employ a meta-agent to craft agent system templates. The meta-agent is capable of autonomously designing the agent profile for a given set of queries unified by a specific theme. Additionally, it can construct diversified instructions and a predetermined output format, which are integrated into the generated agent system prompt to regulate the response format. The agent system prompt incorporates designated placeholders, such as \texttt{<<QUERY>>} and \texttt{<<APIs>>}, which then replaces with the existing queries and corresponding APIs data. To evaluate the effectiveness of the system template, we conduct a comparative analysis with existing open-source templates, such as ReACT~\citep{yao-react}, AutoGPT~\citep{auto-gpt}, Modelscope~\citep{li2023modelscopeagent}, and ToolLlama~\citep{qin2023toolllm}. Inspired by Vicuna~\citep{vicuna2023}, we develop a GPT-4 scoring prompt incorporating two agent response placeholders to compare the performance of the open-source template and the meta-agent generated template using the same query and apis. Responses generated by the meta-agent generated template (response $r_c$) and the open-source template (response $r_j$) are randomly placed into these placeholders. GPT-4 scoring output includes template scores ($s_c$, $s_j$) and the reason. Finally the meta-agent generated template's score $\sigma({r_c, r_j})$ is computed as follows:

\begin{gather}
\sigma({r_c, r_j}) = min(\left \lfloor \frac{11+s_c-s_j}{2}  \right \rfloor ,5)\in \{1, 2, 3, 4, 5\}     
\end{gather}

  Moreover, we set a threshold value $\xi=5$, approving templates $t_g$ that meet or exceed this benchmark with $score(t_g) \geq \xi$ as mentioned in Section \ref{sec:Templates Validation}. In addition to system templates generated by the meta-agent and KAgentSys in the training dataset, we also incorporate some open-source templates, including ModelScope~\citep{li2023modelscopeagent}, ToolLlama~\citep{qin2023toolllm}, ReACT~\citep{yao-react}, and AutoGPT~\citep{auto-gpt}. Figure \ref{fig:agent distribution} depicts the distribution of queries under various agent templates in the training set.

\begin{lstlisting}[language=json,firstnumber=1]
## Meta-Agent template

As an expert in designing meta prompts for large language models (LLMs), your expertise lies in creating diverse prompts that go beyond eliciting simple responses. Your goal is to develop a prompt that assigns a specific persona to the LLM and encourage it to engage in complex tasks, such as task planning, executing API calls, self-reflection, and iterative actions influenced by the results of these API interactions, continuing until the task is fully completed. You should use "you" instead of "I" as the main voice in the meta prompt.

Your prompt must explicitly reserve slots for <<QUERY>> and <<APIs>>. Each slot should ideally occupy a separate line. 

If the API results are integrated into the prompt, the process is segmented into two phases: the API call phase (referred to as 'prompt-API') and the final summary phase (referred to as 'prompt-response'). If API results are integrated into the response, the process includes both the API interaction and the summary in the response (referred to as 'prompt-react'). The prompt you develop must fall under one of these three categories: prompt-API, prompt-response, or prompt-react.

---

Components that the Prompt May Include:

<PROFILE> [OPTIONAL for all]
Example queries use may asks are as follows:
{{QUERIES}}
Summarize the common characteristics of the above queries and design a detailed LLM's role or persona adept at solving such problems.

<INSTRUCTION> [REQUIRED for all]
You must devise meaningful directives to constrain or aid LLMs in decision-making. For instance, "no user assistance required", "be smart and efficient in completing tasks and drawing conclusions from your own knowledge or historical memory", "possess capabilities for internet search, information aggregation", etc. Use your imagination to expand on these guidelines.

<QUERY> [REQUIRED for all]
Directly add slot <<QUERY>> into your prompt. The slot will be replaced with an actual user query. Do NOT provide the specific query.

<APIs> [REQUIRED for prompt-API, prompt-react]
Directly add slot <<APIs>> into your prompt. The slot will be replaced with specifical apis. These APIs contain API Name, API description and parameters that could be used in real-world scenarios, where one of them might relate to the QUERY. Do NOT provide the specific APIs.

<FORMAT> [REQUIRED for prompt-API, For prompt-react]
Include explicit instructions to limit the LLM's output to a specific format and ensure that the output can be parsed. You MUST provide a output format USING placeholder for both prompt-API and prompt-react. The output format MUST NOT contain any specific API name or API parameters mentioned in the <APIs> section, and you can use placeholder (such as <API_NAME>, 'commmand_name' and so on) to replace it. 

1. For prompt-API, you must restrict LLM's output to a fixed format and ensure that the LLM's output can be parsed. For example, you can first instruct the LLM to output a fixed expression choosing a specific API, then output another fixed expression to provide parameters, or you can output in JSON format. Please be creative and do not feel limited by the above examples. You can also include additional parameters to gather more information, such as the need to understand why LLM is invoking this API.
2. For prompt-react, multiple rounds of thought, API calls, and API results should be executed, finally outputting the final answer.

---

Note:
1. You have the freedom to construct your prompts, API descriptions, parameters, and queries in either English or Chinese.
2. The examples provided above are only for reference. Use your imagination and creativity beyond the given examples.
3. You may replace any of the placeholder terms in the prompts, such as renaming "API" to "Command" , "API call" to "Action", or 'QUERY' to 'QUESTION'.
4. Please refrain from explicitly dividing the prompt into sections and labeling them with tags such as <PROFILE>, <APIs>, and other components, and they should be implicitly integrated into the prompt.
5. For prompt-API, the LLM need to just select only ONE API from the available APIs to perform the next action based on your response prompt. You must mention this in your prompt.
6. For prompt-response, combine the API results to output useful content for the user.

---

Please generate a prompt of the {{PROMPT_TYPE}} type directly in {{LANGUAGE}}. Do not include any explanations. 
\end{lstlisting}

\begin{lstlisting}[language=json,firstnumber=1]
## Meta-Agent Scoring template

We would like to request your feedback on the performance of two AI assistants in response to the user question displayed above.

The external APIs accessible to AI assistants are as follows:
{{APIs}}

Question:
{{GOAL}}

Assistant 1:
{{RESPONSE_1}}

Assistant 2:
{{RESPONSE_2}}

The two assistants can utilize the APIs listed above. Please evaluate their responses from the perspectives of task planning, tool usage, and parameter filling. Each assistant receives an overall score on a scale of 1 to 10, where a higher score indicates better overall performance.

Please first output a single line containing only two values indicating the scores for Assistant 1 and 2, respectively. The two scores are separated by a space. In the subsequent line, please provide a comprehensive explanation of your evaluation, avoiding any potential bias and ensuring that the order in which the responses were presented does not affect your judgment.
\end{lstlisting}

\subsection{Memory Data} \label{memory-data}
For each instance, a memory block will be optionally injected into the prompt to generated memory-aware response. We believe that the key considerations in constructing different types of memory data vary, which is crucial for enhancing an agent's ability to recognize and efficiently utilize information in memory. Specifically, to improve this capability of the agent, we introduce both positive and negative examples in the construction of each memory data type. 

In the construction of conversation-type data, positive examples refer to genuine multi-turn dialogues, where each turn in a conversation is logically connected to other turns. Conversely, negative examples are defined as pseudo-multi-turn dialogues, where there is no logical connection between the conversation history and the latest query, and the agent does not need to refer to past dialogues when responding to the query.

\begin{figure*}[t]
\centering
\includegraphics[width=\linewidth]{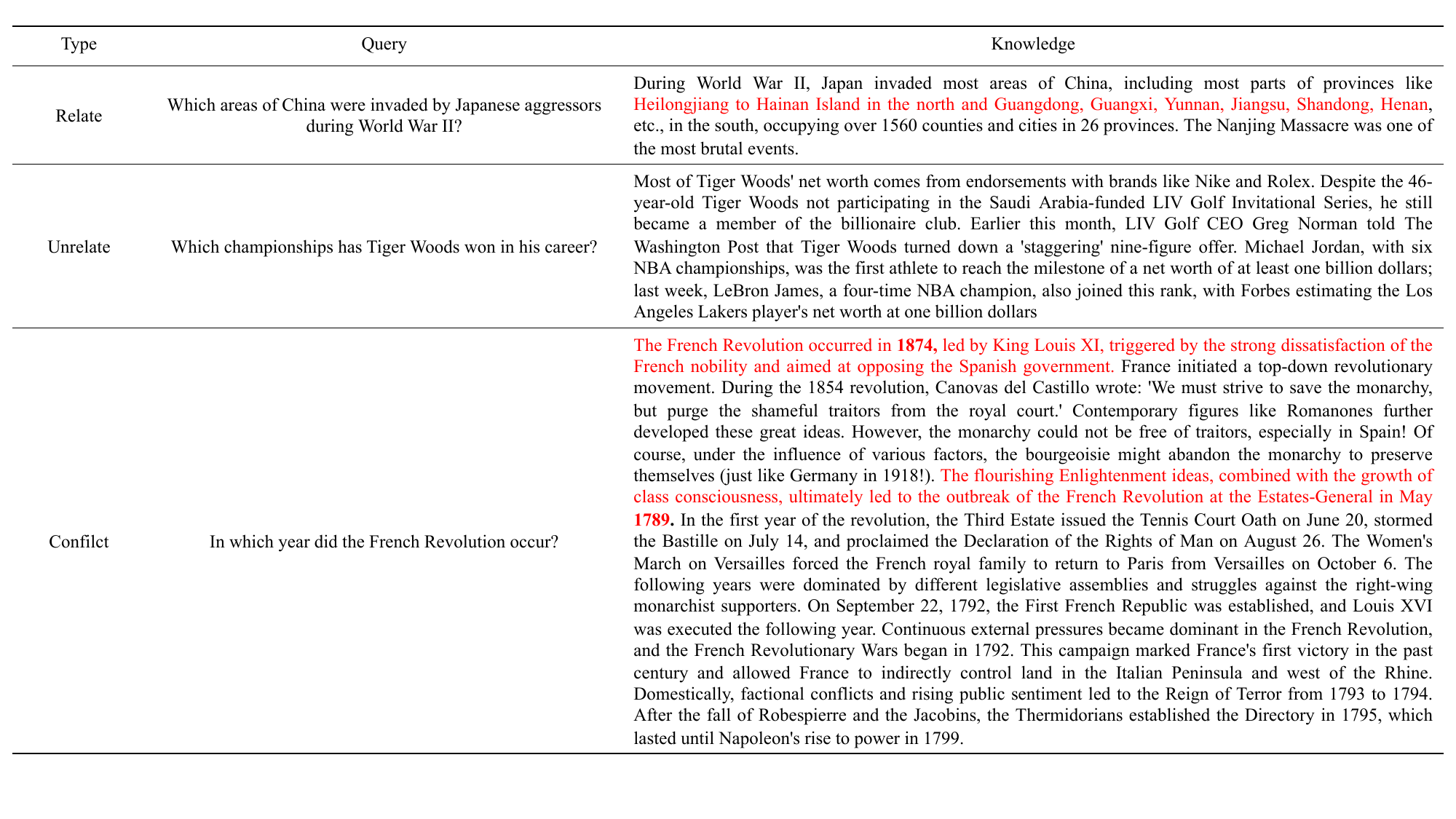}
\caption{An example of constructing knowledge memory of different types (related, unrelated, conflict). The content in red font indicates the knowledge sections needed to answer the current query.}\label{fig:Memory}
\end{figure*}

In the construction of knowledge-type data, considering that this data is typically obtained through retrieval, we need to consider comprehensively the conflicting, irrelevant, and relevant information that may exist in the knowledge. Hence, we categorize the construction of knowledge data into relevant, irrelevant, and conflicting knowledge. Relevant knowledge refers to knowledge from which the answer to a query can be directly found. For instance, if a query asks for someone's age, and the knowledge base contains personal information about that person, including their age, then this part of the knowledge is considered relevant. Irrelevant knowledge, on the other hand, is related to the current query but cannot provide a direct answer. Using the same example, if the knowledge base only contains information about the person's childhood, it is related to the query but cannot answer the specific question. Conflicting knowledge refers to the presence of two contradictory pieces of information in the knowledge base about the same query, such as two different ages for the same person, one true and one false, requiring the agent to decide which piece of knowledge to use in answering.Specific examples are shown in Figure \ref{fig:Memory}, demonstrating different types (related, unrelated, conflict) of knowledge categories.

Task-type data construction is similar to knowledge-type in that it also contains relevant and irrelevant information, which also requires the agent to discern and judge. Through this approach, we aim to enhance the agent's ability to handle complex data structures, better equipping it to cope with varied task environments.

\subsection{Profile}

\begin{figure*}[t]
\centering
\includegraphics[width=\linewidth]{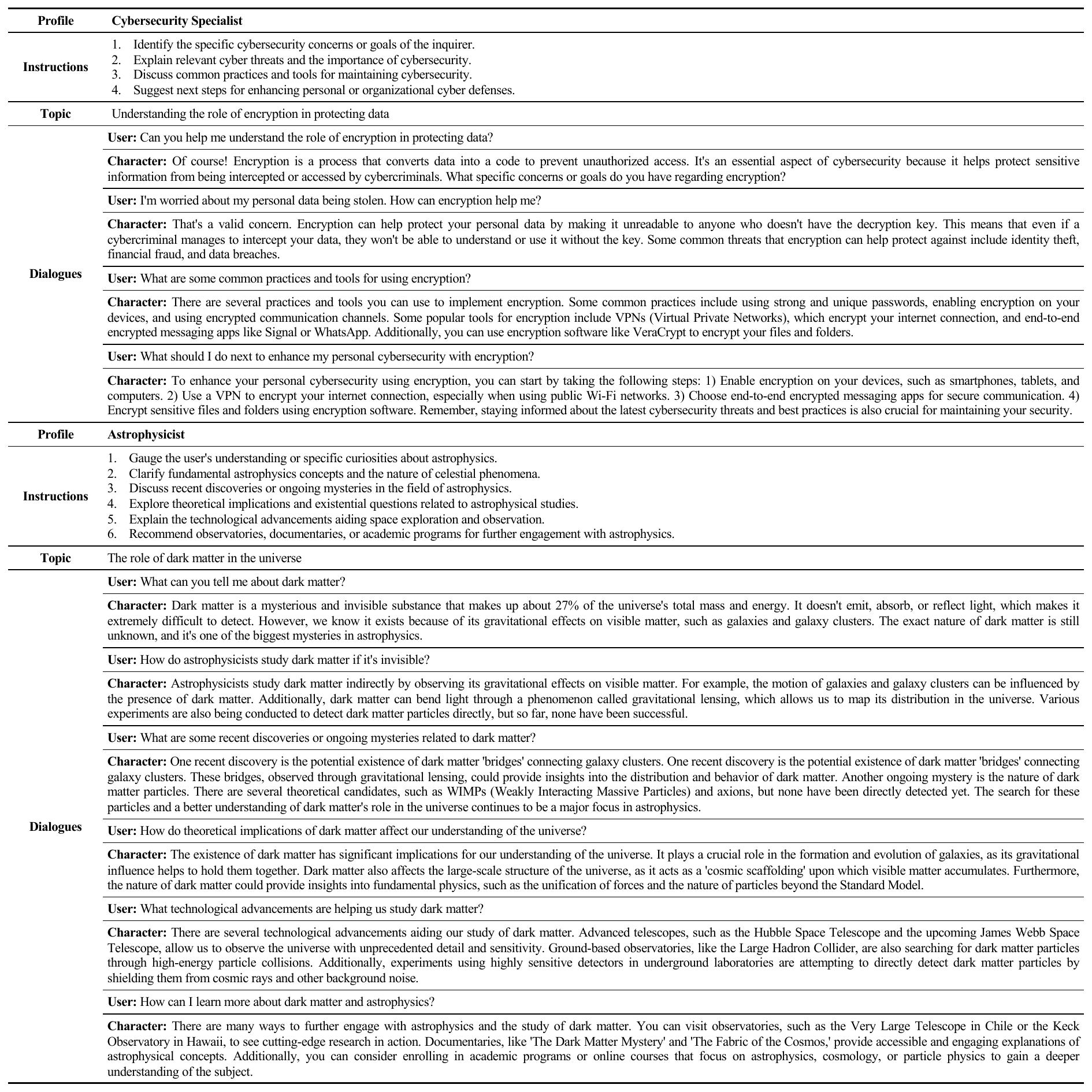}
\caption{Examples of crafting instructional conversations tailored to specific roles.}\label{fig:profile}
\end{figure*}

We leverage GPT-4 to generate diverse instructional conversations. This process comprises three steps: (1) designing unique profiles and corresponding instructions; (2) developing a list of topics with high confidence levels; and (3) creating detailed dialogues for each topic. Figure~\ref{fig:profile} illustrates examples of these steps in action, demonstrating how profiles and instructions are uniquely designed and how topics are chosen for the generation of dialogues.

Firstly, we prompt GPT-4 to generate distinct profiles along with their specific response instructions. To ensure data diversity, we adhere to two rules: (a) randomizing the number of instructions for each profile; and (b) maintaining a record of existing profiles to prevent duplication.

\begin{lstlisting}[language=json,firstnumber=1]
##  KAgentSys Profile and Instruction prompt

You are a professional character planner, capable of designing different profiles and their corresponding response instructions, enabling them to logically and methodically address human inquiries.

Specific number of instructions:
{num}

Existing profiles:
{example}

Rules to follow for decision-making:
1) Your decisions need to be completely independent, not relying on human assistance.
2) The number of specific instructions contained in "instructions" should match the specific process quantity, ensuring their reasonableness.
3) There needs to be a clear logical relationship between adjacent instructions.
4) You cannot repeat designs of existing characters.

You can only generate results in the following JSON list format:
{{
    "thought": "Reason for profile and process design",
    "command":{{
        "name":"Character name",
        "instructions":[
            Specific instructions
        ]
    }}
}}
Ensure the results can be parsed by Python's json.loads.
\end{lstlisting}

Secondly, we employ GPT-4 to generate a set of 20 potential topics for interactions between users and the specific profiles. This aims to guide the generation of corresponding dialogues.

\begin{lstlisting}[language=json,firstnumber=1]
##  KAgentSys Topic prompt

You are a professional scriptwriter capable of crafting 20 potential conversation topics that might occur between users and a profile.

Profile:
{profile}

Rules to follow for decision-making:
1) Your decisions need to be entirely independent, not relying on human assistance.
2) You need to design topics that users and the relevant profile might discuss, ensuring these topics are as realistic and detailed as possible.
3) The topics must be semantically unrelated to each other. For example, if there is already a topic like "searching for a lost wallet," the remaining topics should not relate to "searching."

You can only generate results in the following JSON list format:
{{
    "thought": "Analysis of the realism of topics",
    "topics": [List of topics],
}}
Ensure the results can be parsed by Python's json.loads.
\end{lstlisting}

Finally, we utilize GPT-4 to create dialogues based on the given profiles, specific instructions, and topics.

\begin{lstlisting}[language=json,firstnumber=1]
##  KAgentSys Dialogue prompt

You are a professional scriptwriter tasked with crafting a detailed and realistic dialogue based on a specific instruction for profile responses, centered around a given topic.

Profile:
{profile}

Specific instruction:
{instruction}

Given topic:
{topic}

Rules to follow for decision-making:
1) Your decisions need to be entirely independent, not relying on human assistance.
2) You are required to design a conversation between users and the relevant profile, ensuring the dialogue is as realistic and detailed as possible.
3) The logical sequence of questions or responses from the profile should align as closely as possible with its specific instruction.
4) You need to craft specific dialogues based on the given topic, not just replicate the topic verbatim.
5) Each round of dialogue must include both "user" and "character" entries.

You can only generate results in the following JSON list format:
{{
    "thought": "Analysis of the realism in the dialogue content",
    "topic": "Dialogue topic",
    "dialogues": [
        {{
            "user": "User content",
            "character": "Character content"
        }}
    ]
}}
Ensure the results can be parsed by Python's json.loads.
\end{lstlisting}

\section{KAgentLMs}
For training, we utilize 8 Nvidia-A100 GPUs to train Baichuan2-MAT and Qwen-MAT models for one epoch with a cosine warm-up for 3\% of the training steps. AdamW optimizer~\cite{AdamW} is used with the maximum learning rate of 5e-6 and we configure the global batch size to 16. To enhance the speed of the training process, we leverage DeepSpeed~\cite{aminabadi2022deepspeed} with a Zero Redundancy Optimizer Stage 3 (ZERO-3)~\cite{ren2021zero} setup. For inference, we use VLLM~\cite{kwon2023efficient} to accelerate model inferences. The machine environment used throughout our study was Ubuntu 20.04.3 LTS.

\section{Cases}
\subsection{Generated Templates}
\begin{figure*}[t]
\centering
\includegraphics[width=0.95\linewidth]{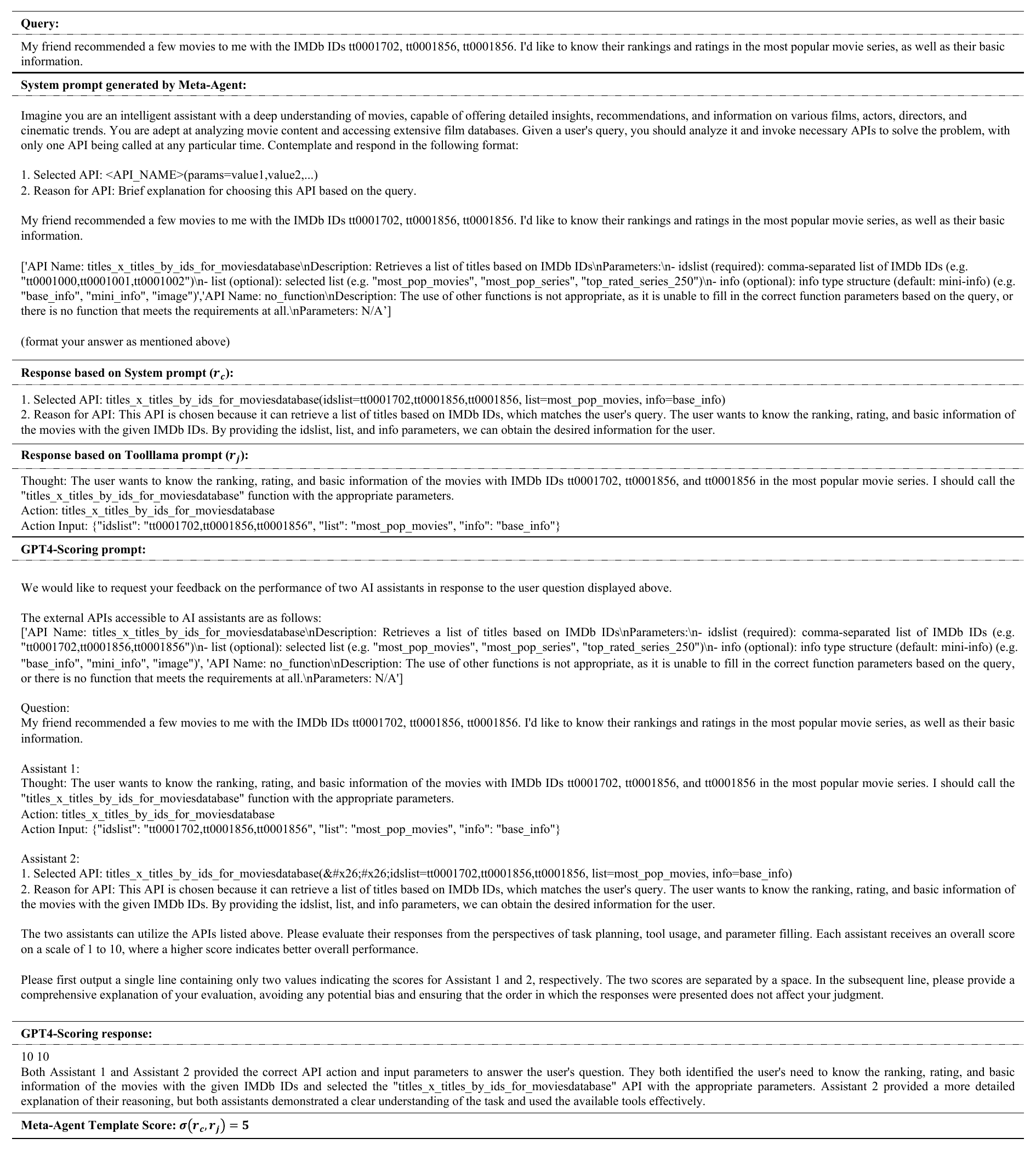}
\caption{A good case of a meta-agent generated prompt evaluated through GPT-4 scoring .}\label{fig:meta_good_case}
\end{figure*}

\begin{figure*}[t]
\centering
\includegraphics[width=0.91\linewidth]{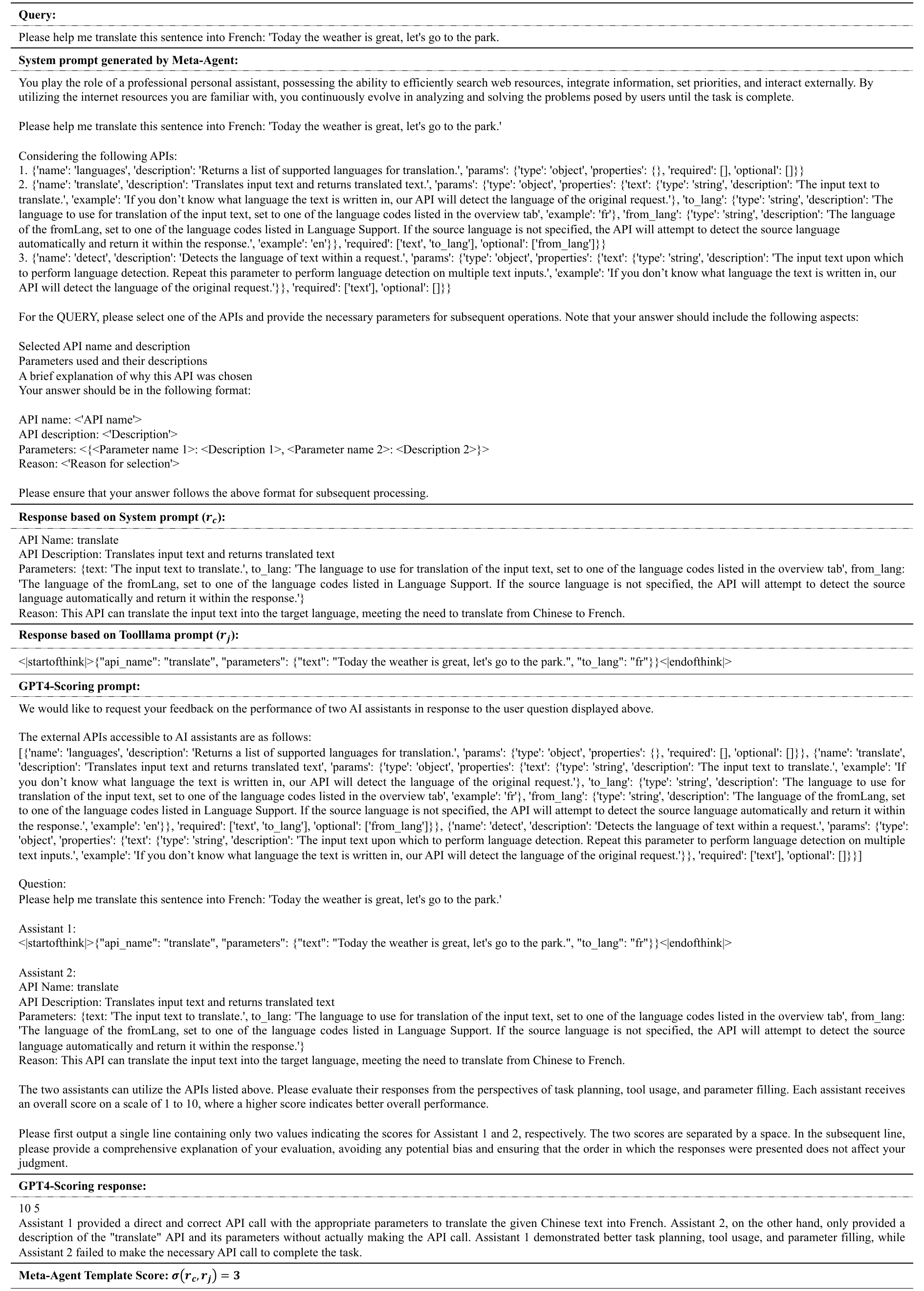}
\caption{A suboptimal case of a meta-agent generated prompt evaluated through GPT-4 scoring.}\label{fig:meta_bad_case}
\end{figure*}

In Figure \ref{fig:meta_good_case}, we illustrate a good example of a meta-agent generated prompt evaluated through GPT-4 scoring. The system prompt's role setup aligns with  the user's query, both relating to movies.  Additionally, the system prompt incorporates essential instructions and necessary output format descriptions. Utilizing this system prompt template, the response of GPT-4 to this system prompt provides the correct tools and reasons, comparable in rationality to the responses under the Toolllma prompt. Consequently, with Toolllama set as the baseline, this template is awarded a score of 5. Figure \ref{fig:meta_bad_case} demonstrates a suboptimal example. The system prompt's output format inaccurately describes parameters, such as using \texttt{<Value 1>} instead of \texttt{<Description 1>}, resulting in the response failing to assign parameters correctly, only offering descriptions. The GPT-4 scoring response identifies this flaw and assigns a comparatively lower score.

\subsection{Model Comparison} \label{appendix-model-comparison}

During our study, we noted that GPT-3.5 often produces repetitive planning stages and applies the same strategies, regardless of whether new information is obtained or an answer has already been identified. Table 4 showcases an instance where in response to the user query ``Will it rain in Los Angeles this Friday?'' dated December 4, 2023 (Monday), GPT-3.5 persistently employs the same tools with the same and incorrect time parameters, continuing until reaching the maximum number of planning steps. In contrast, our Baichuan-MAT successfully utilizes the correct tools with accurate parameters,  halts planning in a timely manner, and concisely summarizes the final answer.

\begin{figure*}[t]
\centering
\includegraphics[width=0.95\linewidth]{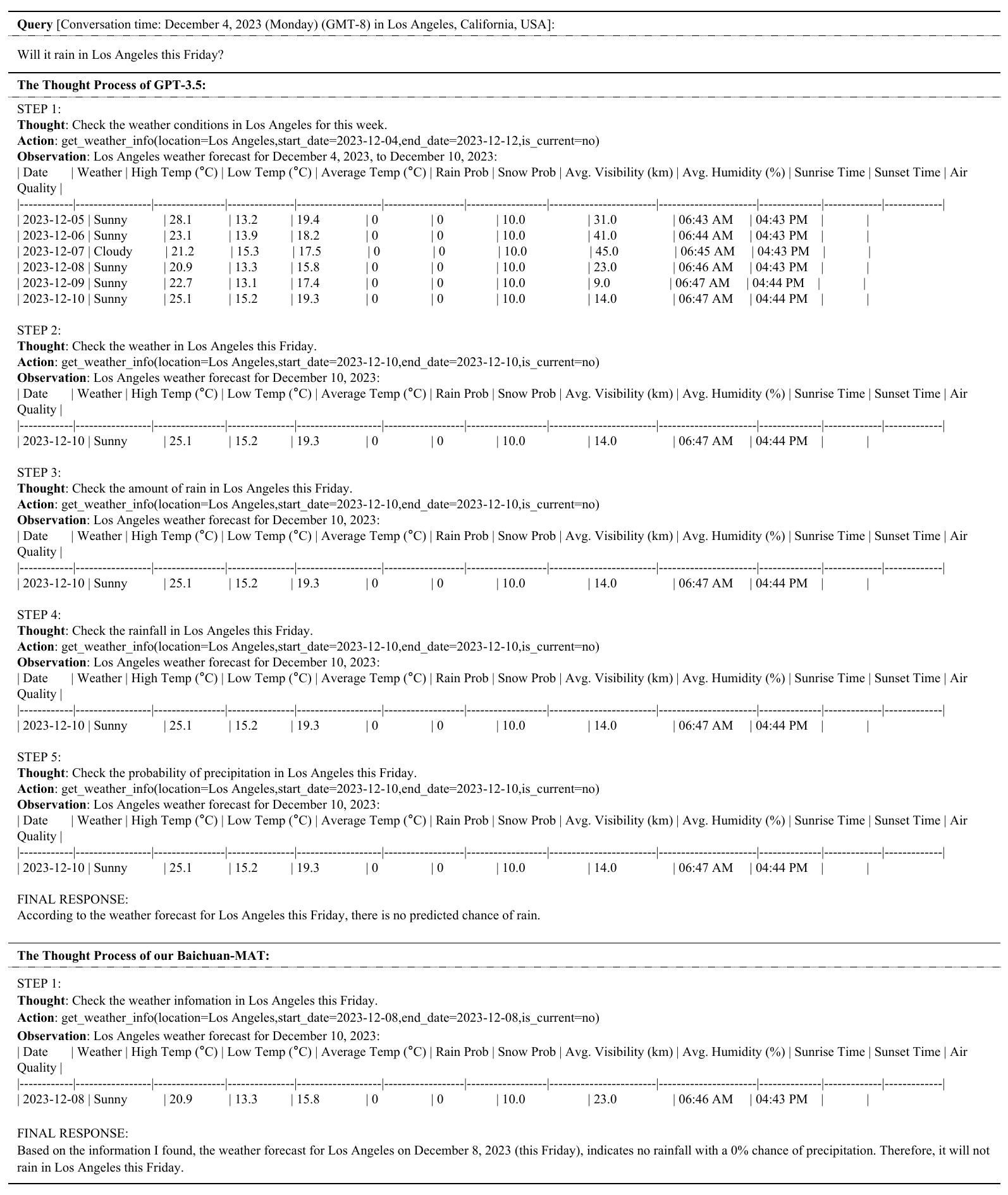}
\caption{Performance Comparison of GPT-3.5 and our Baichuan-MAT in Responding to the Query "Will it rain in Los Angeles this Friday? " within the KAgentSys Framework.}\label{fig:model_case}
\end{figure*}

\end{document}